\documentclass{article}

\usepackage{amsmath,amsfonts}
\usepackage{graphicx}
\usepackage{amssymb}
\usepackage[square, comma, sort&compress, numbers]{natbib}
\usepackage{multirow}
\usepackage[ruled,linesnumbered]{algorithm2e}
\usepackage{amsmath}
\usepackage{pifont}
\usepackage{array}
\usepackage{diagbox}
\usepackage{wrapfig}
\usepackage{setspace}  
\usepackage{wrapfig}
\usepackage{appendix}

\usepackage{authblk}

\usepackage{parskip}[0pt]

\usepackage[preprint]{neurips_2024}

\usepackage[utf8]{inputenc} 
\usepackage[T1]{fontenc}    
\usepackage{hyperref}       
\usepackage{url}            
\usepackage{booktabs}       
\usepackage{amsfonts}       
\usepackage{nicefrac}       
\usepackage{microtype}      
\usepackage{xcolor}         

\title{Advancing Spiking Neural Networks towards Multiscale Spatiotemporal Interaction Learning}

\author{\textbf{Yimeng Shan}$^{1}$, \textbf{Malu Zhang}$^{2}$, \textbf{Rui-Jie Zhu}$^{3}$, \textbf{Xuerui Qiu}$^{2,4}$, \\
\textbf{Jason K. Eshraghian}$^{3}$, \textbf{Haicheng Qu}$^{1}$\thanks{Corresponding author.}\\ 
~\\
$^{1}$Liaoning Technical University, Huludao, Liaoning, China\\
$^{2}$University of Electronic Science and Technology of China, China\\
$^{3}$University of California, Santa Cruz, CA, USA\\
$^{4}$Institute of Automation, Chinese Academy of Sciences, China\\
}

\begin{document}

\maketitle

\begin{abstract}
Recent advancements in neuroscience research have propelled the development of Spiking Neural Networks (SNNs), which not only have the potential to further advance neuroscience research but also serve as an energy-efficient alternative to Artificial Neural Networks (ANNs) due to their spike-driven characteristics. However, previous studies often neglected the multiscale information and its spatiotemporal correlation between event data, leading SNN models to approximate each frame of input events as static images. We hypothesize that this oversimplification significantly contributes to the performance gap between SNNs and traditional ANNs. To address this issue, we have designed a Spiking Multiscale Attention (SMA) module that captures multiscale spatiotemporal interaction information. Furthermore, we developed a regularization method named Attention ZoneOut (AZO), which utilizes spatiotemporal attention weights to reduce the model's generalization error through pseudo-ensemble training. Our approach has achieved state-of-the-art results on mainstream neural morphology datasets. Additionally, we have reached a performance of 77.1\% on the Imagenet-1K dataset using a 104-layer ResNet architecture enhanced with SMA and AZO. This achievement confirms the state-of-the-art performance of SNNs with non-transformer architectures and underscores the effectiveness of our method in bridging the performance gap between SNN models and traditional ANN models. All source code and models are available at \url{https://github.com/Ym-Shan/Spiking_Multiscale_Attention_Arxiv}.

\end{abstract}

\section{Introduction}

The biological brain has long served as a rich source of inspiration for the development of neural networks. By successfully mimicking the complex hierarchical structure of the visual cortex, Artificial Neural Networks (ANNs) have achieved numerous impressive accomplishments~\cite{resnet, yolo}. However, the increasing energy consumption has become a major limitation hindering the further progress of ANNs. In comparison, Spiking Neural Networks (SNNs), which utilize binary spiking signals (0 for no spiking or 1 for spiking), possess inherent low-power characteristics. These low-power attributes of SNNs mainly arise from non-continuous activation~\cite{maass1997networks} and spike-driven properties~\cite{roy2019towards}. The emergence of neuromorphic chips such as Loihi~\cite{davies2018loihi} and TrueNorth~\cite{merolla2014million} is expected to accelerate the widespread adoption of SNNs. The primary goal and ongoing challenge in SNNs research is to simultaneously draw inspiration from both high-performance deep learning models and the mechanisms of the biological brain in order to achieve brain-inspires intelligence.

In the initial stages of SNNs' development, researchers encountered challenges in training algorithms. Drawing inspiration from synaptic plasticity in neuroscience and the backpropagation algorithm in deep learning, they proposed training methods such as STDP unsupervised learning~\cite{tao2023new}, ANN2SNN~\cite{deng2021optimal, wu2021progressive}, and STBP~\cite{wu2018spatio}. Building upon this foundation, some researchers incorporated elements from deep learning, such as VGG~\cite{spikingvgg} and ResNet~\cite{sew, ms} architectures, into SNNs or optimized loss functions to enhance model performance. Additionally, motivated by insights from neuroscience, some researchers introduced attention mechanisms into SNNs~\cite{ma,tcja} or designed neuron simulation models that mimic the real properties of the biological brain, aiming to achieve more brain-inspires characteristics.

\begin{figure}[ht]
  \centering 
  \includegraphics[width=1\textwidth]{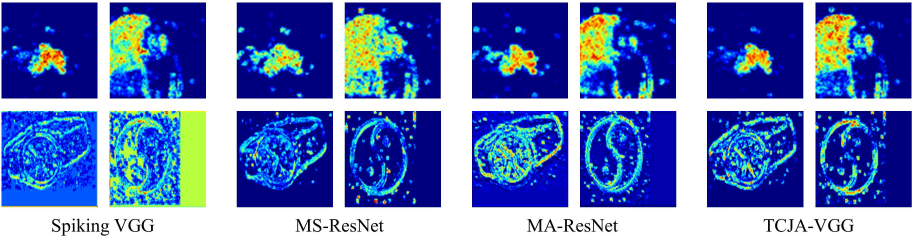}
  \caption{Displaying Learning Patterns of Several Mainstream Models. All four images depict attention heat maps based on the Spiking Firing Rate (SFR), with red indicating high activation and blue representing low spiking activity.}
  \label{show_patten} 
\end{figure}

However, most prior studies have not delineated discrepancies in resolution and morphology among sample features within mainstream datasets. Consequently, the majority of these works have concentrated on devising comprehensive network architectures rather than considering multiscale structures in their designs. Furthermore, with the exception of TCJA~\cite{tcja}, none of the works have taken into account the spatiotemporal correlations. Visualizing the spike dynamics of the bottleneck neurons in mainstream networks, as depicted in Fig.~\ref{show_patten}, we analyze their learning patterns. The results indicate that SNN models, which overlook multiscale information and spatiotemporal correlations, exhibit recognition patterns for event data akin to ANN networks recognizing static images. Moreover, SNN models relying solely on spatiotemporal correlations fail to alter this recognition pattern. We posit that this oversight has led to contemporary SNNs forfeiting some brain-inspires characteristics.

Therefore, we propose a Spiking Multiscale Attention (SMA) module to introduce multiscale representation learning into the SNN domain, while simultaneously leveraging spatiotemporal correlations to compute attention weights to address this issue. This method enhances the model's ability to extract multiscale features while altering the learning pattern of SNNs, enabling SNN models to better balance local and global features. It is worth noting that we attribute the performance gap between SNNs and ANNs to the insufficient utilization of spatiotemporal correlation information by SNN models. To further mitigate this limitation, we devised an Attention Zoneout (AZO) regularization method. This method enhances model generalization by replacing information at spatiotemporal weak points of hidden units with information from previous time step. Unlike dropout, the AZO regularization method involves substitution rather than deletion, facilitating smoother propagation of gradient and state information through time.
The primary contributions of this paper are summarized as follows:

\begin{itemize}
\setlength{\itemindent}{0pt}
\item
We observe a neglect in previous works towards multiscale information and spatiotemporal correlation, and identify that this leads to the loss of brain-inspired features in SNN models, consequently leading to erroneous learning patterns.
\item
We extend the SE module~\cite{se} to multiple scales and propose the SMA module based on it to introduce multiscale representation learning into SNNs, while utilizing spatiotemporal correlation information to compute attention weights for balancing local and global information, thereby steering SNN models towards a more brain-like learning pattern. To the best of our knowledge, this marks the first attempt to integrate multiscale representation learning into SNNs.
\item
To further leverage spatiotemporal correlation information, we introduce an AZO regularization method leveraging spatiotemporal attention weights. This method employs the previous hidden unit value as noise to replace irrelevant information, thereby training a pseudo ensemble to enhance the robustness and generalization of the model.
\item
We demonstrate the effectiveness of our proposed approach by achieving state-of-the-art accuracy on three mainstream datasets and Imagenet-1K. Through comprehensive visualization analysis, we provide evidence that the method of simultaneously leveraging multiscale information and spatiotemporal correlation can indeed induce SNNs to adopt a more brain-inspires learning pattern.
\end{itemize}

\section{Related works}

\textbf{Attention Mechanism} has emerged as a key factor in improving deep learning model performance, complementing depth, width, and cardinality. As traditional model structure design reaches a bottleneck, developing more effective attention mechanisms to focus on crucial features has become essential. In the SNN realm, Yao et al. introduced the TA module~\cite{ta} as the inaugural attention mechanism emphasizing feature importance in the time dimension. Subsequently, the same team devised the MA module~\cite{ma} to simultaneously highlight feature significance across time, spatial, and channel dimensions. From a different perspective, Zhu et al. proposed the tcja module~\cite{tcja}, a portable and efficient solution to better concentrate on the characteristics of the time and channel dimensions, considering their interrelation. Shan et al. pioneered the integration of discrete spiking signals into attention-based decision-making processes~\cite{orrc}. Considering the need for lightweight design and the utilization of multiscale spatiotemporal correlation information, we developed the MSE module and the SMA module based on it, which transform the model's learning pattern at minimal cost.

\textbf{Multiscale Representation Learning} plays a pivotal role in various computer vision applications, including image classification~\cite{mul_image_classification_1, mul_image_classification_2, mul_image_classification_3}, object detection~\cite{mul_target_detection_1, mul_target_detection_2, mul_target_detection_3}, image segmentation~\cite{mul_img_segmentation_1, mul_img_segmentation_2}, and image fusion~\cite{mul_fusion_1, mul_fusion_2}. Given the diverse shapes and resolutions of natural objects, multiscale approaches enable the identification of relevant features~\cite{TAI_review}. With the advancement of deep learning, it has been observed that the depth of a network correlates positively with its feature representation capability~\cite{resnet, densenet}. To enhance this capability, researchers have integrated multiscale representation learning into deep learning frameworks by designing multiscale convolution structures~\cite{li2019selective}, incorporating pyramid structures~\cite{mtcnn, dssd, fpn}, formulating multiscale loss functions~\cite{refinenet, pyramid}, and developing multiscale attention mechanisms~\cite{zhang2021multiscale, zheng2022msa}. In this work, we broaden the scope of multiscale representation learning to SNNs by introducing a multiscale attention module.

\textbf{Regularization} reduces a model's generalization error, typically dominated by variance, by balancing added bias against reduced variance. Data regularization methods like Cutout~\cite{cutout}, Mixup~\cite{zhang2017mixup}, and CutMix~\cite{yun2019cutmix} apply input image transformations to enhance model robustness. Structural regularization includes Dropout~\cite{fast_dropout_training}, adapted for specific tasks like Zoneout~\cite{krueger2016zoneout} for sequences and Dropblock~\cite{ghiasi2018dropblock} for segmentation, improving performance by selectively retaining unit values. Due to the approximate nature of dropout, researchers have explored optimal locations; Maxdropout~\cite{do2021maxdropout} selects based on neuron activation, while Autodropout~\cite{pham2021autodropout} chooses based on training insights. Our AZO regularization method utilizes SMA attention weights for selection, replacing irrelevant information with the hidden unit values from previous timesteps as noise, thereby training a pseudo-ensemble to enhance robustness and generalizability.

\section{Method}
\subsection{Leaky Integrate-and-Fire Neuron Model}
The Leaky Integrate-and-Fire (LIF) neuron~\cite{maass1997networks} stands as one of the predominant neuron models within SNNs, esteemed for its balanced performance and adherence to biological principles, is uniformly adopted in this work. Within neural networks, neurons serve as fundamental computational units. Upon receiving transmitted spiking signals, LIF neurons initiate an integration process. Upon reaching a threshold membrane potential, neurons emit pulses and reset their membrane potentials. This dynamic process is encapsulated by the subthreshold dynamics model~\cite{roy2019towards}

\vspace{-12pt}
\begin{equation}
\label{eq1}
\tau \frac{{d\boldsymbol{V}(t)}}{{dt}} =  - (\boldsymbol{V}(t) - {\boldsymbol{V}_{reset}}) + \boldsymbol{I}(t),
\end{equation}
\vspace{-12pt}

where $\tau$ represents a time constant, $\boldsymbol{V}(t)$ denotes the membrane potential of the postsynaptic neuron, and $\boldsymbol{I}(t)$ signifies the input gathered from presynaptic neurons. Additionally, $\boldsymbol{V}_{\text{reset}}$ denotes the reset potential, which is established subsequent to the activation of the output spiking.To facilitate training and description, we adopt the displayed iteration version of the subthreshold dynamics model~\cite{neftci2019surrogate}.

\vspace{-12pt}
\begin{equation}
\label{eq2}
\boldsymbol{U}_t^n = \boldsymbol{H}_{t - 1}^n + \frac{1}{\tau }(\boldsymbol{I}_{t - 1}^n - (\boldsymbol{H}_{t - 1}^n -\boldsymbol{U}_{reset})),
\end{equation}
\vspace{-10pt}

\vspace{-10pt}
\begin{equation}
\label{eq3}
\boldsymbol{S}_t^n = \Theta (\boldsymbol{U}_t^n - {\boldsymbol{U}_{threshold}}),
\end{equation}
\vspace{-10pt}

\vspace{-10pt}
\begin{equation}
\label{eq4}
\boldsymbol{H}_t^n = \boldsymbol{U}_t^n(1 - \boldsymbol{S}_t^n),
\end{equation}
\vspace{-12pt}

at each layer indexed by $n$ and time step $t$, the membrane potential $\boldsymbol{U}$ of a neuron is denoted as $\boldsymbol{U}_{\text{t}}^{^\text{n}}$. The parameter $\tau$ signifies a time constant, and $\boldsymbol{S}$ represents a binary spiking tensor. $\boldsymbol{I}$ denotes the neuron's input, while $\Theta$ represents the Heaviside step function. $\boldsymbol{H}$ symbolizes the hidden state, $\boldsymbol{U}_{\text{reset}}$ refers to the reset potential of the neuron following a spike, and $\boldsymbol{U}_{\text{threshold}}$ indicates the discharge threshold of the neuron.

\begin{figure}[ht]
  \centering 
  \includegraphics[width=1\textwidth]{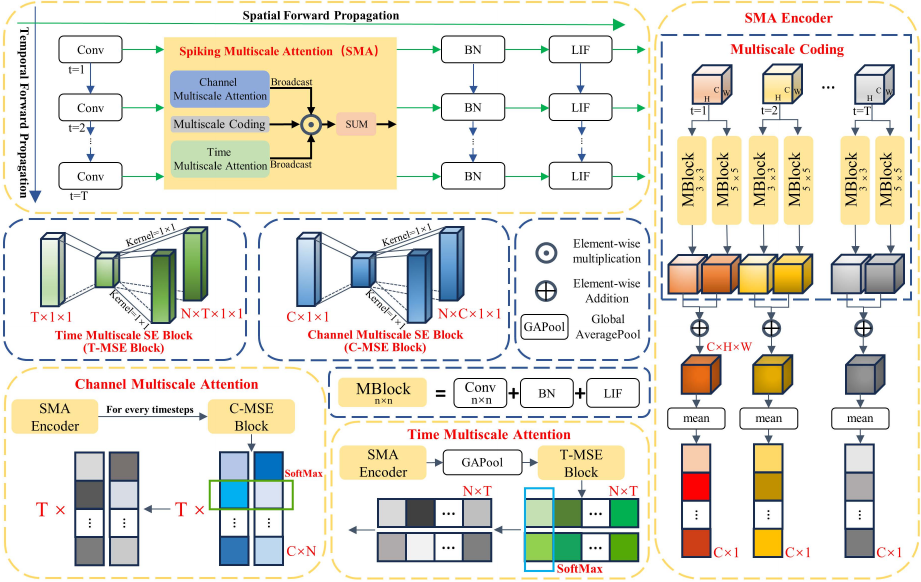}
  \caption{The overview of Spiking Multiscale Attention (SMA) module. In the figure, the schematic diagram of the encoder is shown on the right side, the schematic diagram of the decoder is displayed in the lower-left corner, and the schematic diagram of the Multiscale SE (MSE) block module is positioned in the center.}
  \label{fig2} 
\end{figure}

\subsection{Spiking Multiscale Attention for SNNs}

The overall structure of the SMA module is illustrated in Fig.~\ref{fig2}. Its primary objective is to enhance the model's ability to extract target features with diverse resolutions, shapes, and states by utilizing spatiotemporal correlation information. This approach helps balance the significance of global and local features, thereby transforming the model's learning pattern. The temporal and spatial information inherent in event data can often be intertwined, posing additional challenges in extracting multiscale features. Hence, instead of directly conducting multiscale feature extraction within the encoder, we opt to perform multiscale coding beforehand to enhance feature representation. The encoder's architecture is delineated in the right portion of Fig.~\ref{fig2}. Given an input event sequence $\boldsymbol{X} = [ \cdots ,{\boldsymbol{X}_t}, \cdots ] \in {\mathbb{R}^{T \times C \times H \times W}}$, the encoder of the SMA module across $\boldsymbol{N}$ scales can be represented as

\vspace{-12pt}
\begin{equation}
\label{eq5}
\boldsymbol{E}(x,k) = \delta (BN(Conv2d(x,k))),
\end{equation}
\vspace{-10pt}

\vspace{-10pt}
\begin{equation}
\label{eq6}
\begin{array}{*{20}{c}}
{\boldsymbol{M}_t^n = E({\boldsymbol{X}_t},\boldsymbol{K}_t^n),}&{n \in \boldsymbol{N},t \in \boldsymbol{T},\boldsymbol{M}_t^n \in {\mathbb{R}^{C \times H \times W}}}
\end{array}
\end{equation}
\vspace{-10pt}

\vspace{-14pt}
\begin{equation}
\label{eq7}
\begin{array}{*{20}{c}}
\begin{array}{*{20}{c}}
{{\boldsymbol{Y}_t} = \sum\limits_{n = 1}^N {\frac{{\boldsymbol{M}_t^n}}{\boldsymbol{N}},} }&{{\boldsymbol{Y}_t} \in {\mathbb{R}^{C \times H \times W}}}
\end{array}
\end{array}
\end{equation}
\vspace{-12pt}

where $\boldsymbol{E}(x,k)$ represents the multiscale coding process and $\delta$ denotes the $ReLU$ function. However, when analyzing the mechanism of SMA, we substitute LIF neurons for $\delta$ (we have conducted ample experiments in Sec.~\ref{sec:comparison} of the appendix demonstrating that LIF does not notably compromise model accuracy compared to $ReLU$). $\boldsymbol{C}onv2d(x,k)$ indicates that the 2D convolution operation is performed on $\boldsymbol{X}$ using the convolution kernel $\boldsymbol{K}$, and ${\boldsymbol{M}_t} = [ \cdots ,\boldsymbol{M}_t^n, \cdots ] \in {\mathbb{R}^{N \times C \times H \times W}}$ represents the multiscale coding result of the t-th frame of the input event.

The decoder structure is illustrated in the lower-left portion of Fig.~\ref{fig2}. It is responsible for processing the multiscale data $\boldsymbol{Y} = [ \cdots ,{\boldsymbol{Y}_t}, \cdots ] \in {\mathbb{R}^{T \times C \times H \times W}}$, enriched with characteristics output from the encoder. Specifically, it calculates the attention weights for the temporal and channel dimensions as follows:

\vspace{-12pt}
\begin{equation}
\label{eq8}
f_\alpha ^n(\boldsymbol{X}) = \boldsymbol{C}onv2d(\boldsymbol{K}_{E,\alpha }^n,\delta (\boldsymbol{C}onv2d({\boldsymbol{K}_{S,\alpha }},\boldsymbol{X}))),
\end{equation}
\vspace{-10pt}

\vspace{-10pt}
\begin{equation}
\label{eq9}
f_\beta ^n(\boldsymbol{X}) = \boldsymbol{C}onv2d(\boldsymbol{K}_{E,\beta }^n,\delta (\boldsymbol{C}onv2d({\boldsymbol{K}_{S,\beta }},\boldsymbol{X}))),
\end{equation}
\vspace{-10pt}

\vspace{-10pt}
\begin{equation}
\label{eq10}
\begin{array}{*{20}{c}}
{{\boldsymbol{W}_\alpha } = \eta({f_\alpha }(\boldsymbol{A}vgpool({\boldsymbol{Y}_t}))),}&{{\boldsymbol{W}_\alpha } \in {\mathbb{R}^{N \times T \times 1}}}
\end{array}
\end{equation}
\vspace{-10pt}

\vspace{-10pt}
\begin{equation}
\label{eq11}
\begin{array}{*{20}{c}}
{{\boldsymbol{W}_{\beta ,t}} = \eta({f_\beta }({\boldsymbol{Y}_t})),}&{{\boldsymbol{W}_{\beta ,t}} \in {\mathbb{R}^{N \times C \times 1}}}
\end{array}
\end{equation}
\vspace{-12pt}

where $f_\alpha ^n(\boldsymbol{X})$ and $f_\beta ^n(\boldsymbol{X})$ respectively describe the roles of the MSE (T-MSE) module in the time dimension and the MSE (C-MSE) module in the channel dimension. ${\boldsymbol{K}_{S,\alpha }}$ and ${\boldsymbol{K}_{S,\beta }}$ represent the Squeeze convolution kernel ($1 \times 1$) in the T-MSE module and C-MSE module, respectively. $\boldsymbol{K}_{E,\alpha }^n$ and $\boldsymbol{K}_{E,\beta }^n$ represent the Excitation convolution kernel ($1 \times 1$) in the n-th scale, respectively. $\eta$ represents the $SoftMax$ operation.

To compute attention weights in the time dimension, we adhere to the conventional methodology: initially conducting global average pooling on the input data, followed by computing attention weights to derive weights sized $T \times 1$. To comprehensively leverage the interplay between multiscale information and spatiotemporal interactions, attention weights are computed for all input channels at each timestep. It is noteworthy that upon completing the calculation, $SoftMax$ operations are performed on the weights within the time and channel dimensions correspondingly, in the scale dimension, to mitigate the adverse effects of large (small) values on the model. Consequently, ${\boldsymbol{W}_\beta } = [ \cdots ,{\boldsymbol{W}_{c,t}}, \cdots ] \in {\mathbb{R}^{N \times T \times C \times H \times W}}$ are obtained.Finally, apply the attention weights to the multiscale encoding result $\boldsymbol{M} = [ \cdots ,{\boldsymbol{M}_t}, \cdots ] \in {\mathbb{R}^{N \times T \times C \times H \times W}}$ of the input event stream and aggregate them along the scale dimension:

\vspace{-12pt}
\begin{equation}
\label{eq12}
\begin{array}{*{20}{c}}
{\boldsymbol{Z} = \boldsymbol{S}um(\boldsymbol{M} \times {\boldsymbol{W}_\alpha } \times {\boldsymbol{W}_\beta }).}&{\boldsymbol{Z} \in {\mathbb{R}^{T \times C \times H \times W}}}
\end{array}
\end{equation}
\vspace{-12pt}

\subsection{Attention Zoneout for SMA}

\begin{algorithm}
    \caption{Attention Zoneout}
    \label{alg1}
    \KwIn{Output of SMA module decoder : $\boldsymbol{Z}$; Time attention weight : ${\boldsymbol{W}_\alpha }$; Channel attention weight : ${\boldsymbol{W}_\beta }$; Number of timesteps executing AZO : ${\delta _t}$; Number of channels executing AZO : ${\delta _c}$}
    \KwOut{Z after AZO execution : $\boldsymbol{R}$}
    $ \boldsymbol{R} = \boldsymbol{Z} $\\
    Find the indices of the ${\delta _t}$ smallest values in array ${\boldsymbol{W}_\alpha }$ : $\boldsymbol{H} \in {\mathbb{N}^{{\delta _t} \times 1}}$ \\
    $ \boldsymbol{H} = \boldsymbol{S}ort(\boldsymbol{H}, Ascending) $\\
    \For{each $ i = 0,1, \cdots ,\boldsymbol{H} $}
        {
            Find the indices of the ${\delta _c}$ smallest values in array ${\boldsymbol{W}_{\beta ,i}}$ : $ {\boldsymbol{P}_i} \in {\mathbb{N}^{{\delta _c} \times 1}} $
        }
    \For{each $ i = 0,1, \cdots ,\boldsymbol{H} $}
        {
            \For{each $ j = 0,1, \cdots ,{\boldsymbol{P}_i} $}
                {
                    $\begin{array}{*{20}{c}} 
                    {\boldsymbol{R}[i][j] = \boldsymbol{Z}[i - 1][j]}&{if(j \ne 0)} 
                    \end{array}$
                }
        }
    return $ \boldsymbol{R} $
    
\end{algorithm}

Zhu et al. were the first to explore the correlation between temporal and spatial information in event data within the domain of SNNs~\cite{tcja}. However, most mainstream SNN methods overlook this spatiotemporal correlation information. To address this, we propose a regularization method called Attention Zoneout (AZO), which utilizes temporal and channel attention weights, as described in Algorithm~\ref{alg1}, to leverage these valuable pieces of information.

Similar to Zoneout~\cite{krueger2016zoneout}, AZO trains pseudo-ensembles by adding noise to hidden units, thereby enhancing generalization ability. The key difference is that the location of noise in AZO is determined by attention weights rather than being random, which helps mitigate the impact of irrelevant features while introducing noise, thus facilitating the network's convergence to the global optimum. Additionally, while Zoneout directly adds noise in the spatial dimension, AZO applies noise in the channel dimension because, at the time of AZO operation, all features have already been aggregated onto the channel dimension, resulting in a spatial dimension size of $1 \times 1$. Consequently, each operation applied to the channel dimension is equivalently projected onto the spatial dimension. In Sec.~\ref{sec:AZO Hyp} of the appendix, we present an ablation study to elucidate the selection process of the hyperparameters ${\delta_t}$ and ${\delta_c}$ in AZO and provide additional experiments to validate its efficacy.

\section{Experiments}
Even with the use of spiking coding, the spatiotemporal interaction information in static image datasets remains limited. Therefore, we only evaluated the classification performance of our proposed SMA-SNN and SMA-AZO-SNN architectures on three prominent neuromorphic datasets (DVS128 Gesture~\cite{amir2017low}, CIFAR10-DVS~\cite{li2017cifar10}, and N-Caltech101~\cite{orchard2015converting}) and the ImageNet-1K dataset~\cite{deng2009imagenet}, as elaborated in Sec.~\ref{sec:exp/compa}. Preceding this assessment, we provided a comprehensive exposition of our attention position determination and scale quantity selection process in Sec.~\ref{sec:exp/abl}, a crucial aspect we regard as essential for a generic attention module. All network structures and hyperparameter settings utilized in the experiment are detailed in Sec.~\ref{sec:model} of appendix.


\subsection{Ablation study}
\label{sec:exp/abl}

\begin{wrapfigure}[23]{r}{0.6\textwidth}
\vspace{-10pt}
  \includegraphics[width=0.6\textwidth]{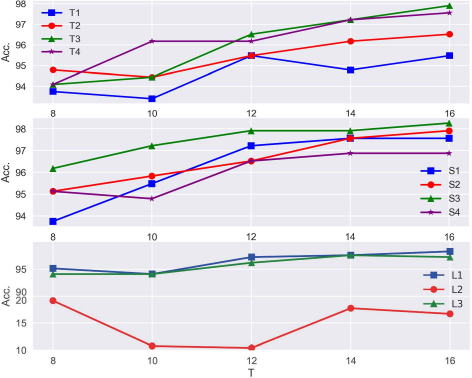}
  \caption{Ablation study of different SMA positions based on DVS128 Gesture. Inspired by previous work~\cite{ta, ma}, we placed SMA behind convolution layer in the first two groups of experiments.}
  \label{location} 
\end{wrapfigure}

Building upon the foundation established by prior research, it has been observed that the classification performance of SNN models on the DVS128 Gesture dataset~\cite{amir2017low} can be effectively extrapolated to other prominent datasets~\cite{ta, ma, tcja, sew}. Consequently, our ablation studies will be systematically conducted utilizing the DVS128 Gesture dataset as the primary benchmark.

\textbf{SMA Position.} Incorporating the plug-and-play attention module into mainstream network architectures necessitates careful consideration of its insertion points and quantity. Initially, we address the unique characteristics of the encoding module (the first module of SNNs) and categorize the potential roles of SMA within the network as follows: \textbf{T1, absence of SMA; T2, exclusive addition of SMA to the encoding module; T3, integration of SMA throughout the network except for the encoding module; T4, comprehensive inclusion of SMA across the entire network.} As depicted in Fig.~\ref{location}, T3 consistently outperforms the other configurations across nearly all simulation timesteps, particularly at the prevalent timestep of 16.

Building upon the aforementioned findings, we further subdivide the potential optimal positions for SMA within the network as follows: \textbf{S1, incorporating SMA solely in odd-numbered blocks excluding the encoding module; S2, integrating SMA exclusively in even-numbered blocks; S3, embedding SMA only in the initial half of the network, but excluding the coding module; S4, inserting SMA solely in the latter half of the network.} As illustrated in Fig.~\ref{location}, the S3 configuration demonstrates superior performance compared to other positions across all simulated timesteps. Additionally, we investigated the influence of SMA within the conventional SNN module (Conv-BN-Neuron). The potential insertion points for SMA are: \textbf{L1, immediately following the convolution layer; L2, following the BN layer; L3, after the neuron layer.} The experimental findings demonstrate that within the module, the placement of the SMA module at position L1 exhibits a slight advantage over its placement at L3, while positioning it at L2 results in the network's failure to converge. This phenomenon is hypothesized to stem from the SMA module's inability to accurately extract significance from normalized data, consequently leading to the disappearance of the network's gradient.

\begin{wraptable}{r}{7.5cm}
\vspace{-0.7cm}
\caption{Various Scale Quantities of SMA. The filter sizes used for sampling at each scale are progressively 1, 3, 5, 7, 9, and 11. The experimental results are derived from ten identical experiments.}
\label{scale_battle}
\centering
\begin{tabular}{@{}ccc@{}}
\toprule
\textbf{Scale}     & \textbf{Acc(\%)}     & \textbf{Inference Overhead(s)} \\ \midrule
No SMA & 96.52       & 18.98                 \\
2         & 96.52($\pm$0.34) & 21.99($\pm$0.98)            \\
3         & 97.21($\pm$0.69) & 23.02($\pm$0.07)           \\
4         & 97.91($\pm$0.34) & 27.44($\pm$0.24)           \\
5         & 97.73($\pm$0.17) & 33.72($\pm$0.2)           \\
6         & 97.56($\pm$0.34) & 42.43($\pm$0.06)           \\ \bottomrule
\end{tabular}
\end{wraptable}

\textbf{Scale Quantities.} Paying attention to various receptive fields of differing sizes is crucial for SMA. The results in Tab.~\ref{scale_battle} demonstrate that the SMA module, integrating downsampling and attention mechanisms utilizing filters of sizes 1, 3, 5, and 7 across four distinct scales, yields the most pronounced and consistent effects. Furthermore, this configuration introduces only marginal additional reasoning time, thereby facilitating efficient processing for individual events without a significant increase in computational load.

\subsection{Comparison with the State-of-the-Art}
\label{sec:exp/compa}

\begin{table}[]
\caption{Evaluation on ImageNet-1K. In inference, the default resolution of the input crops is $224 \times 224$. The experimental input crops marked with * are enlarged to $288 \times 288$.}
\label{imagenet}
\centering
\begin{tabular}{@{}cccc@{}}
\toprule
\textbf{Work}                                      & \textbf{Architecture}               & \textbf{Timestep} & \textbf{Top-1 Acc.(\%)} \\ \midrule
\multirow{4}{*}{SEW ResNet~\cite{sew}}               & SEW-ResNet-34              & 4        & 67.04          \\
                                          & SEW-ResNet-50              & 4        & 67.78          \\
                                          & SEW-ResNet-101             & 4        & 68.76          \\
                                          & SEW-ResNet-152             & 4        & 69.26          \\ \cmidrule(l){2-4} 
\multirow{4}{*}{MS ResNet~\cite{ms}}                & MS-ResNet-18               & 6        & 63.10          \\
                                          & MS-ResNet-34               & 6        & 69.42          \\
                                          & MS-ResNet-104              & 5        & 74.21          \\
                                          & MS-ResNet-104*              & 5        & 76.02          \\ \cmidrule(l){2-4} 
\multirow{3}{*}{MA-ResNet~\cite{ma}}                & MA-MS-ResNet-18            & 1        & 63.97          \\
                                          & MA-MS-ResNet-34            & 1        & 69.15          \\
                                          & MA-MS-ResNet-104*           & 4        & \textbf{77.08}          \\ \cmidrule(l){2-4} 
Spiking ResNet~\cite{hu2021spiking}                            & ResNet-50                  & 350      & 72.75          \\
Hybrid training~\cite{rathi2020enabling}                           & ResNet-34                  & 250      & 61.48          \\
TET~\cite{deng2022temporal}                                       & SEW-ResNet-34              & 4        & 68.00          \\
tdBN~\cite{zheng2021going}                                      & Spiking-ResNet-34          & 6        & 63.72          \\
Spike-Norm~\cite{spikingvgg}                                & ResNet34                   & 2000     & 65.47          \\
QCFS~\cite{bu2023optimal}                                      & VGG-16                     & 64       & 72.85          \\
Fast-SNN~\cite{hu2023fast}                                  & VGG-16                     & 7        & 72.95          \\    \midrule
ResNet~\cite{ms}                                    & Res-CNN-104                & 1        & 76.87          \\ \midrule
\multirow{2}{*}{Spikformer~\cite{zhou2022spikformer}}               & Spiking Transformer-10-512 & 4        & 73.68          \\
                                          & Spiking Transformer-8-768*  & 4        & 74.81          \\ \cmidrule(l){2-4} 
\multirow{2}{*}{Spike-driven Transformer~\cite{yao2024spike}} & Spiking Transformer-10-512 & 4        & 74.66          \\
                                          & Spiking Transformer-8-768  & 4        & \textbf{77.07}          \\ \cmidrule(l){2-4}

\multirow{2}{*}{Meta-SpikeFormer~\cite{yao2024spike222}} & Meta-SpikeFormer & 4        & \textbf{77.20}          \\
                                          & Meta-SpikeFormer*  & 4        & \textbf{80.00}          \\ \midrule

\multirow{3}{*}{\textbf{SMA-ResNet(Ours)}}               & SMA-MS-ResNet-18           & 6        & \textbf{68.46}                \\
                                          & SMA-MS-ResNet-34           & 6        & \textbf{70.19}               \\
                                          & SMA-AZO-MS-ResNet-104*          & 5        & \textbf{77.05}              \\ \bottomrule
\end{tabular}
\end{table}

\textbf{Experimental results on Imagenet-1K.} We adopted the same approach as Yao et al~\cite{ma}, simply integrating the SMA module into MS-ResNet~\cite{ms} to evaluate its performance. Tab.~\ref{imagenet} presents the experimental results. The integration of SMA alone significantly improved model accuracy, evidenced by enhancements of 5.36\% in the 18-layer ResNet and 0.77\% in the 34-layer ResNet configurations. Furthermore, the combined application of SMA and AZO resulted in an additional accuracy increase of 1.03\% in the 104-layer structure. Comparison with previous studies indicates that our approach achieves state-of-the-art accuracy in traditional convolutional-based SNNs and is also competitive with some models based on the Transformer architecture.

\begin{table}[]
\caption{The comparison between the proposed methods and existing SOTA techniques on three
mainstream neuromorphic datasets.}
\label{neurodatasets}
\centering
\begin{tabular}{@{}c@{\hspace{0cm}}cp{0.8cm}<{\centering}ccccc@{}}
\toprule
\multirow{2}{*}{\textbf{Work}}                                              & \multirow{2}{*}{\textbf{Spike-driven}} & \multicolumn{2}{c}{\textbf{DVS128 Gesture}} & \multicolumn{2}{c}{\textbf{CIFAR10-DVS}} & \multicolumn{2}{c}{\textbf{N-Caltech101}} \\ \cmidrule(l){3-8} 
                                                                   &                               & \textbf{T}               & \textbf{Acc}              & \textbf{T}             & \textbf{Acc}             & \textbf{T}              & \textbf{Acc}             \\ \midrule
PLIF~\cite{fang2021incorporating}                                                               & \ding{51}                  & 20              & 97.6             & 20            & 74.8            & -              & -               \\
Spikformer~\cite{zhou2022spikformer}                                                         & \ding{55}                        & \textbf{16}              & \textbf{98.3}             & 16            & 80.9            & -              & -               \\
tdBN~\cite{zheng2021going}                                                               & \ding{55}                        & 40              & 96.9             & 10            & 67.8            & -              & -               \\
SEW-ResNet~\cite{sew}                                                         & \ding{55}                        & 16              & 97.9             & 16            & 74.4            & -              & -               \\
TA-SNN~\cite{ta}                                                             & \ding{55}                       & 60              & 98.6             & 10            & 72.0            & -              & -               \\
HATS~\cite{sironi2018hats}                                                               & N/A                           &  -               & -                & N/A           & 52.4            & N/A            & 64.2            \\
DART~\cite{ramesh2019dart}                                                               & N/A                           & -               & -                & N/A           & 65.8            & N/A            & 66.8            \\
SALT~\cite{kim2021optimizing}                                                               & \ding{55}                              & -               & -                & 20            & 67.1            & 20             & 55.0            \\
TCJA-SNN~\cite{tcja}                                                           & \ding{55}                              & \textbf{20}              & \textbf{99.0}             & 10            & 80.7            & 14             & 78.5            \\
\begin{tabular}[c]{@{}c@{}}Spike-driven \\ Transformer~\cite{yao2024spike}\end{tabular} &  \ding{51}                            & \textbf{16}              & \textbf{99.3}             & 16            & 80.0            & -              & -               \\ \midrule
\textbf{SMA-VGG(Ours)}                                                            & \ding{51}                    & 16              & 98.3             & 10            & \textbf{83.1}           & 14             & \textbf{83.7}           \\
\textbf{SMA-AZO-VGG(Ours)}                                                        & \ding{51}                    & \textbf{16}              & \textbf{98.6}             & 10            & \textbf{84.0}              & 14                & \textbf{84.6}                \\ \bottomrule
\end{tabular}
\end{table}

\textbf{Experimental results on mainstream neuromorphic datasets.} As Tab.~\ref{neurodatasets} illustrates, the integration of SMA into the VGG network allows for SOTA performance on the CIFAR10-DVS and N-Caltech101 datasets. The accuracy is further and steadily improved through the application of the AZO regularization method. Our approach also resulted in a significant 6.1\% increase in accuracy on the N-Caltech101 dataset. Moreover, on the DVS128 Gesture dataset, our method achieved equivalent results to those of non-transformer architectures with the same timesteps.

\subsection{Study of Learning Patterns}

\begin{figure}[ht]
  \centering 
  \includegraphics[width=1\textwidth]{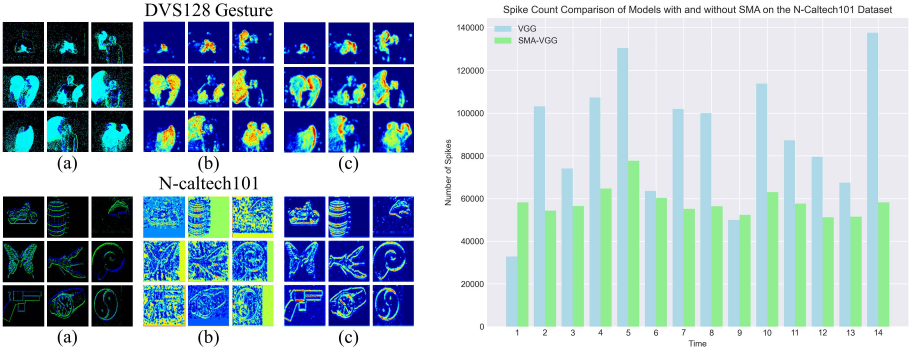}
  \caption{Visualization of typical sample input frames and their attentional heatmaps: (a) shows the input frame; (b) and (c) display attention heat maps based on the Spiking Firing Rate (SFR), where red indicates high and blue shows low spiking activation. Heat map (b) is from the Spiking-VGG8 model and (c) from the SMA-SNN model. All heat maps are from the first convolutional layer of each model, except the coding layer. A spike count comparison is shown on the right.}
  \label{heatmap} 
\end{figure}

\textbf{SMA Changes SNNs Learning Patterns.} Observations of the attention heatmaps depicted in Fig.~\ref{show_patten} and \ref{heatmap}(b) reveal that traditional SNN decision-making approaches treat each frame of event data as a static image. This strategy leads to an overemphasis on global features and a neglect of local features, thereby impeding the utilization of temporal information. However, as depicted in Fig.~\ref{heatmap}(c), the incorporation of the SMA module balances the importance of local and global features by integrating multiscale and spatiotemporal correlation information. This shift in the model's decision-making mode towards recognizing the relative positions and dynamic information of crucial joints enhances the utilization of spatiotemporal correlation information between event frames.


The N-Caltech101 dataset has more complex features and more noise. A closer examination of Fig.~\ref{heatmap} reveals that SNN models lacking SMA tend to allocate unnecessary attention to the background while neglecting local features. We identify this as the primary reason for the underperformance of traditional SNN models on the N-Caltech101 dataset. Upon integrating SMA, the SMA-SNN model effectively mitigates the issue of background overemphasis observed in traditional SNNs, redirecting attention more precisely towards relevant local features. This enhancement is believed to be the primary factor contributing to the improved performance of the SMA-SNN model on the N-Caltech101 dataset. Additionally, as depicted in Fig.~\ref{heatmap} right, SMA significantly reduces the SFR of the SNN model by minimizing unnecessary attention to the background, thereby promising greater energy savings for event-driven SNNs.

\subsection{Study on the correlation between SMA and multiscale information in samples}

\begin{wrapfigure}[16]{r}{0.6\textwidth}
\vspace{-2pt}
  \includegraphics[width=0.6\textwidth]{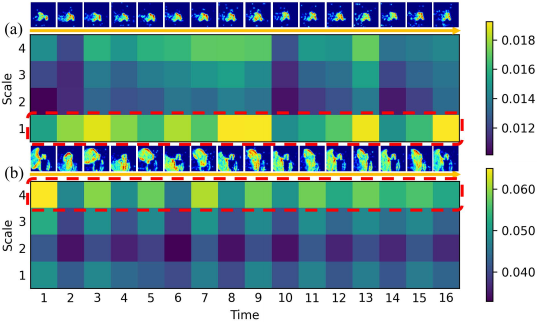}
  \caption{Scale importance between different types of samples. }
  \label{scale} 
\end{wrapfigure}

\textbf{Focus on multiscale features.} The filters of sizes 1, 3, 5, and 7 in Fig.~\ref{scale} are employed for downsampling across the four scales. Research conducted on this figure suggests that in the decision-making process, samples characterized by smaller key features (such as fingers and wrists) exhibit a preference for filters of smaller size, whereas samples with larger key features (such as arms and trajectories) show a predilection for filters of larger size. This observation underscores the intrinsic capability of the SMA module to enhance model performance in datasets with a diverse range of feature sizes.

\section{Conclusion}

In our observations of previous work, we found that SNN models that ignore multiscale information and spatiotemporal correlation information exhibit recognition patterns for event data similar to ANN models recognizing static images. Additionally, visualizations of the heatmaps from previous work indicate that solely relying on spatiotemporal correlation information cannot fundamentally alter the recognition patterns of SNNs. Therefore, we introduced multiscale spatiotemporal interaction learning into SNNs by designing the Spiking Multiscale Attention (SMA) module to alter the learning patterns of SNNs. Experiments demonstrate that SMA effectively utilizes both multiscale information and spatiotemporal correlation information to balance the importance of global and local features in the samples, thereby changing the learning patterns of SNNs. We further utilize spatiotemporal correlation information by proposing an AZO regularization method. This method replaces the information at spatiotemporal weak points with the information from the corresponding spatial locations at the previous time step to train pseudo-ensembles, effectively reducing the model's generalization error. SMA and AZO achieve state-of-the-art accuracy on CIFAR10-DVS~(84.0\%) and N-Caltech101~(84.6\%) datasets, while also attaining non-Transformer architecture state-of-the-art accuracy on the ImageNet-1k dataset~(77.1\%), underscoring their efficacy. Therefore, our work highlights the importance of multiscale information and spatiotemporal correlation information in SNNs, advancing SNNs one step closer to the decision-making processes of the human brain.

    \bibliographystyle{unsrt}
    \bibliography{ref}

\begin{thebibliography}{10}

\bibitem{resnet}
Christian Szegedy, Wei Liu, Yangqing Jia, Pierre Sermanet, Scott Reed, Dragomir Anguelov, Dumitru Erhan, Vincent Vanhoucke, and Andrew Rabinovich.
\newblock Going deeper with convolutions.
\newblock In {\em Proceedings of the IEEE conference on computer vision and pattern recognition}, pages 1--9, 2015.

\bibitem{yolo}
Joseph Redmon, Santosh Divvala, Ross Girshick, and Ali Farhadi.
\newblock You only look once: Unified, real-time object detection.
\newblock In {\em Proceedings of the IEEE conference on computer vision and pattern recognition}, pages 779--788, 2016.

\bibitem{maass1997networks}
Wolfgang Maass.
\newblock Networks of spiking neurons: the third generation of neural network models.
\newblock {\em Neural networks}, 10(9):1659--1671, 1997.

\bibitem{roy2019towards}
Kaushik Roy, Akhilesh Jaiswal, and Priyadarshini Panda.
\newblock Towards spike-based machine intelligence with neuromorphic computing.
\newblock {\em Nature}, 575(7784):607--617, 2019.

\bibitem{davies2018loihi}
Mike Davies, Narayan Srinivasa, Tsung-Han Lin, Gautham Chinya, Yongqiang Cao, Sri~Harsha Choday, Georgios Dimou, Prasad Joshi, Nabil Imam, Shweta Jain, et~al.
\newblock Loihi: A neuromorphic manycore processor with on-chip learning.
\newblock {\em Ieee Micro}, 38(1):82--99, 2018.

\bibitem{merolla2014million}
Paul~A Merolla, John~V Arthur, Rodrigo Alvarez-Icaza, Andrew~S Cassidy, Jun Sawada, Filipp Akopyan, Bryan~L Jackson, Nabil Imam, Chen Guo, Yutaka Nakamura, et~al.
\newblock A million spiking-neuron integrated circuit with a scalable communication network and interface.
\newblock {\em Science}, 345(6197):668--673, 2014.

\bibitem{tao2023new}
Tuomin Tao, Da~Li, Hanzhi Ma, Yan Li, Shurun Tan, En-xiao Liu, Jose Schutt-Aine, and Er-Ping Li.
\newblock A new pre-conditioned stdp rule and its hardware implementation in neuromorphic crossbar array.
\newblock {\em Neurocomputing}, 557:126682, 2023.

\bibitem{deng2021optimal}
Shikuang Deng and Shi Gu.
\newblock Optimal conversion of conventional artificial neural networks to spiking neural networks.
\newblock {\em arXiv preprint arXiv:2103.00476}, 2021.

\bibitem{wu2021progressive}
Jibin Wu, Chenglin Xu, Xiao Han, Daquan Zhou, Malu Zhang, Haizhou Li, and Kay~Chen Tan.
\newblock Progressive tandem learning for pattern recognition with deep spiking neural networks.
\newblock {\em IEEE Transactions on Pattern Analysis and Machine Intelligence}, 44(11):7824--7840, 2021.

\bibitem{wu2018spatio}
Yujie Wu, Lei Deng, Guoqi Li, and Luping Shi.
\newblock Spatio-temporal backpropagation for training high-performance spiking neural networks.
\newblock {\em Frontiers in neuroscience}, 12:323875, 2018.

\bibitem{spikingvgg}
Abhronil Sengupta, Yuting Ye, Robert Wang, and Kaushik Roy.
\newblock Going deeper in spiking neural networks: Vgg and residual architectures.
\newblock {\em Frontiers in neuroscience}, 13:425055, 2019.

\bibitem{sew}
Wei Fang, Zhaofei Yu, Yanqi Chen, Tiejun Huang, Timoth{\'e}e Masquelier, and Yonghong Tian.
\newblock Deep residual learning in spiking neural networks.
\newblock {\em Advances in Neural Information Processing Systems}, 34:21056--21069, 2021.

\bibitem{ms}
Yifan Hu, Lei Deng, Yujie Wu, Man Yao, and Guoqi Li.
\newblock Advancing spiking neural networks toward deep residual learning.
\newblock {\em IEEE Transactions on Neural Networks and Learning Systems}, 2024.

\bibitem{ma}
Man Yao, Guangshe Zhao, Hengyu Zhang, Yifan Hu, Lei Deng, Yonghong Tian, Bo~Xu, and Guoqi Li.
\newblock Attention spiking neural networks.
\newblock {\em IEEE transactions on pattern analysis and machine intelligence}, 2023.

\bibitem{tcja}
Rui-Jie Zhu, Qihang Zhao, Tianjing Zhang, Haoyu Deng, Yule Duan, Malu Zhang, and Liang-Jian Deng.
\newblock Tcja-snn: Temporal-channel joint attention for spiking neural networks.
\newblock {\em arXiv preprint arXiv:2206.10177}, 2022.

\bibitem{se}
Jie Hu, Li~Shen, and Gang Sun.
\newblock Squeeze-and-excitation networks.
\newblock In {\em Proceedings of the IEEE conference on computer vision and pattern recognition}, pages 7132--7141, 2018.

\bibitem{ta}
Man Yao, Huanhuan Gao, Guangshe Zhao, Dingheng Wang, Yihan Lin, Zhaoxu Yang, and Guoqi Li.
\newblock Temporal-wise attention spiking neural networks for event streams classification.
\newblock In {\em Proceedings of the IEEE/CVF International Conference on Computer Vision}, pages 10221--10230, 2021.

\bibitem{orrc}
Yimeng Shan, Xuerui Qiu, Rui-jie Zhu, Ruike Li, Meng Wang, and Haicheng Qu.
\newblock Or residual connection achieving comparable accuracy to add residual connection in deep residual spiking neural networks.
\newblock {\em arXiv preprint arXiv:2311.06570}, 2023.

\bibitem{mul_image_classification_1}
Zhiqiang Gong, Ping Zhong, Yang Yu, Weidong Hu, and Shutao Li.
\newblock A cnn with multiscale convolution and diversified metric for hyperspectral image classification.
\newblock {\em IEEE Transactions on Geoscience and Remote Sensing}, 57(6):3599--3618, 2019.

\bibitem{mul_image_classification_2}
Chun-Fu~Richard Chen, Quanfu Fan, and Rameswar Panda.
\newblock Crossvit: Cross-attention multi-scale vision transformer for image classification.
\newblock In {\em Proceedings of the IEEE/CVF international conference on computer vision}, pages 357--366, 2021.

\bibitem{mul_image_classification_3}
Gao Huang, Danlu Chen, Tianhong Li, Felix Wu, Laurens Van Der~Maaten, and Kilian~Q Weinberger.
\newblock Multi-scale dense networks for resource efficient image classification.
\newblock {\em arXiv preprint arXiv:1703.09844}, 2017.

\bibitem{mul_target_detection_1}
Jinhui Han, Kun Liang, Bo~Zhou, Xinying Zhu, Jie Zhao, and Linlin Zhao.
\newblock Infrared small target detection utilizing the multiscale relative local contrast measure.
\newblock {\em IEEE Geoscience and Remote Sensing Letters}, 15(4):612--616, 2018.

\bibitem{mul_target_detection_2}
Puti Yan, Runze Hou, Xuguang Duan, Chengfei Yue, Xin Wang, and Xibin Cao.
\newblock Stdmanet: Spatio-temporal differential multiscale attention network for small moving infrared target detection.
\newblock {\em IEEE transactions on geoscience and remote sensing}, 61:1--16, 2023.

\bibitem{mul_target_detection_3}
Yantao Wei, Xinge You, and Hong Li.
\newblock Multiscale patch-based contrast measure for small infrared target detection.
\newblock {\em Pattern Recognition}, 58:216--226, 2016.

\bibitem{mul_img_segmentation_1}
Eitan Sharon, Achi Brandt, and Ronen Basri.
\newblock Fast multiscale image segmentation.
\newblock In {\em Proceedings IEEE Conference on Computer Vision and Pattern Recognition. CVPR 2000 (Cat. No. PR00662)}, volume~1, pages 70--77. IEEE, 2000.

\bibitem{mul_img_segmentation_2}
Koen~L. Vincken, Andre S.~E. Koster, and Max~A. Viergever.
\newblock Probabilistic multiscale image segmentation.
\newblock {\em IEEE Transactions on Pattern Analysis and Machine Intelligence}, 19(2):109--120, 1997.

\bibitem{mul_fusion_1}
Seong~G Kong, Jingu Heo, Faysal Boughorbel, Yue Zheng, Besma~R Abidi, Andreas Koschan, Mingzhong Yi, and Mongi~A Abidi.
\newblock Multiscale fusion of visible and thermal ir images for illumination-invariant face recognition.
\newblock {\em International Journal of Computer Vision}, 71:215--233, 2007.

\bibitem{mul_fusion_2}
A~Ben~Hamza, Yun He, Hamid Krim, and Alan Willsky.
\newblock A multiscale approach to pixel-level image fusion.
\newblock {\em Integrated Computer-Aided Engineering}, 12(2):135--146, 2005.

\bibitem{TAI_review}
Licheng Jiao, Jie Gao, Xu~Liu, Fang Liu, Shuyuan Yang, and Biao Hou.
\newblock Multiscale representation learning for image classification: A survey.
\newblock {\em IEEE Transactions on Artificial Intelligence}, 4(1):23--43, 2021.

\bibitem{densenet}
Gao Huang, Zhuang Liu, Laurens Van Der~Maaten, and Kilian~Q Weinberger.
\newblock Densely connected convolutional networks.
\newblock In {\em Proceedings of the IEEE conference on computer vision and pattern recognition}, pages 4700--4708, 2017.

\bibitem{li2019selective}
Xiang Li, Wenhai Wang, Xiaolin Hu, and Jian Yang.
\newblock Selective kernel networks.
\newblock In {\em Proceedings of the IEEE/CVF conference on computer vision and pattern recognition}, pages 510--519, 2019.

\bibitem{mtcnn}
Kaipeng Zhang, Zhanpeng Zhang, Zhifeng Li, and Yu~Qiao.
\newblock Joint face detection and alignment using multitask cascaded convolutional networks.
\newblock {\em IEEE signal processing letters}, 23(10):1499--1503, 2016.

\bibitem{dssd}
Cheng-Yang Fu, Wei Liu, Ananth Ranga, Ambrish Tyagi, and Alexander~C Berg.
\newblock Dssd: Deconvolutional single shot detector.
\newblock {\em arXiv preprint arXiv:1701.06659}, 2017.

\bibitem{fpn}
Tsung-Yi Lin, Piotr Doll{\'a}r, Ross Girshick, Kaiming He, Bharath Hariharan, and Serge Belongie.
\newblock Feature pyramid networks for object detection.
\newblock In {\em Proceedings of the IEEE conference on computer vision and pattern recognition}, pages 2117--2125, 2017.

\bibitem{refinenet}
Guosheng Lin, Anton Milan, Chunhua Shen, and Ian Reid.
\newblock Refinenet: Multi-path refinement networks for high-resolution semantic segmentation.
\newblock In {\em Proceedings of the IEEE conference on computer vision and pattern recognition}, pages 1925--1934, 2017.

\bibitem{pyramid}
Hengshuang Zhao, Jianping Shi, Xiaojuan Qi, Xiaogang Wang, and Jiaya Jia.
\newblock Pyramid scene parsing network.
\newblock In {\em Proceedings of the IEEE conference on computer vision and pattern recognition}, pages 2881--2890, 2017.

\bibitem{zhang2021multiscale}
Tao Zhang, Xiao-Qing Zheng, and Ming-Xin Liu.
\newblock Multiscale attention-based lstm for ship motion prediction.
\newblock {\em Ocean Engineering}, 230:109066, 2021.

\bibitem{zheng2022msa}
Linxin Zheng, Guobao Xiao, Ziwei Shi, Shiping Wang, and Jiayi Ma.
\newblock Msa-net: Establishing reliable correspondences by multiscale attention network.
\newblock {\em IEEE Transactions on Image Processing}, 31:4598--4608, 2022.

\bibitem{cutout}
Terrance DeVries and Graham~W Taylor.
\newblock Improved regularization of convolutional neural networks with cutout.
\newblock {\em arXiv preprint arXiv:1708.04552}, 2017.

\bibitem{zhang2017mixup}
Hongyi Zhang, Moustapha Cisse, Yann~N Dauphin, and David Lopez-Paz.
\newblock mixup: Beyond empirical risk minimization.
\newblock {\em arXiv preprint arXiv:1710.09412}, 2017.

\bibitem{yun2019cutmix}
Sangdoo Yun, Dongyoon Han, Seong~Joon Oh, Sanghyuk Chun, Junsuk Choe, and Youngjoon Yoo.
\newblock Cutmix: Regularization strategy to train strong classifiers with localizable features.
\newblock In {\em Proceedings of the IEEE/CVF international conference on computer vision}, pages 6023--6032, 2019.

\bibitem{fast_dropout_training}
Sida Wang and Christopher Manning.
\newblock Fast dropout training.
\newblock In {\em international conference on machine learning}, pages 118--126. PMLR, 2013.

\bibitem{krueger2016zoneout}
David Krueger, Tegan Maharaj, J{\'a}nos Kram{\'a}r, Mohammad Pezeshki, Nicolas Ballas, Nan~Rosemary Ke, Anirudh Goyal, Yoshua Bengio, Aaron Courville, and Chris Pal.
\newblock Zoneout: Regularizing rnns by randomly preserving hidden activations.
\newblock {\em arXiv preprint arXiv:1606.01305}, 2016.

\bibitem{ghiasi2018dropblock}
Golnaz Ghiasi, Tsung-Yi Lin, and Quoc~V Le.
\newblock Dropblock: A regularization method for convolutional networks.
\newblock {\em Advances in neural information processing systems}, 31, 2018.

\bibitem{do2021maxdropout}
Claudio Filipi~Goncalves do~Santos, Danilo Colombo, Mateus Roder, and Jo{\~a}o~Paulo Papa.
\newblock Maxdropout: deep neural network regularization based on maximum output values.
\newblock In {\em 2020 25th International Conference on Pattern Recognition (ICPR)}, pages 2671--2676. IEEE, 2021.

\bibitem{pham2021autodropout}
Hieu Pham and Quoc Le.
\newblock Autodropout: Learning dropout patterns to regularize deep networks.
\newblock In {\em Proceedings of the AAAI Conference on Artificial Intelligence}, volume~35, pages 9351--9359, 2021.

\bibitem{neftci2019surrogate}
Emre~O Neftci, Hesham Mostafa, and Friedemann Zenke.
\newblock Surrogate gradient learning in spiking neural networks: Bringing the power of gradient-based optimization to spiking neural networks.
\newblock {\em IEEE Signal Processing Magazine}, 36(6):51--63, 2019.

\bibitem{amir2017low}
Arnon Amir, Brian Taba, David Berg, Timothy Melano, Jeffrey McKinstry, Carmelo Di~Nolfo, Tapan Nayak, Alexander Andreopoulos, Guillaume Garreau, Marcela Mendoza, et~al.
\newblock A low power, fully event-based gesture recognition system.
\newblock In {\em Proceedings of the IEEE conference on computer vision and pattern recognition}, pages 7243--7252, 2017.

\bibitem{li2017cifar10}
Hongmin Li, Hanchao Liu, Xiangyang Ji, Guoqi Li, and Luping Shi.
\newblock Cifar10-dvs: an event-stream dataset for object classification.
\newblock {\em Frontiers in neuroscience}, 11:244131, 2017.

\bibitem{orchard2015converting}
Garrick Orchard, Ajinkya Jayawant, Gregory~K Cohen, and Nitish Thakor.
\newblock Converting static image datasets to spiking neuromorphic datasets using saccades.
\newblock {\em Frontiers in neuroscience}, 9:159859, 2015.

\bibitem{deng2009imagenet}
Jia Deng, Wei Dong, Richard Socher, Li-Jia Li, Kai Li, and Li~Fei-Fei.
\newblock Imagenet: A large-scale hierarchical image database.
\newblock In {\em 2009 IEEE conference on computer vision and pattern recognition}, pages 248--255. Ieee, 2009.

\bibitem{hu2021spiking}
Yangfan Hu, Huajin Tang, and Gang Pan.
\newblock Spiking deep residual networks.
\newblock {\em IEEE Transactions on Neural Networks and Learning Systems}, 34(8):5200--5205, 2021.

\bibitem{rathi2020enabling}
Nitin Rathi, Gopalakrishnan Srinivasan, Priyadarshini Panda, and Kaushik Roy.
\newblock Enabling deep spiking neural networks with hybrid conversion and spike timing dependent backpropagation.
\newblock {\em arXiv preprint arXiv:2005.01807}, 2020.

\bibitem{deng2022temporal}
Shikuang Deng, Yuhang Li, Shanghang Zhang, and Shi Gu.
\newblock Temporal efficient training of spiking neural network via gradient re-weighting.
\newblock {\em arXiv preprint arXiv:2202.11946}, 2022.

\bibitem{zheng2021going}
Hanle Zheng, Yujie Wu, Lei Deng, Yifan Hu, and Guoqi Li.
\newblock Going deeper with directly-trained larger spiking neural networks.
\newblock In {\em Proceedings of the AAAI conference on artificial intelligence}, volume~35, pages 11062--11070, 2021.

\bibitem{bu2023optimal}
Tong Bu, Wei Fang, Jianhao Ding, PengLin Dai, Zhaofei Yu, and Tiejun Huang.
\newblock Optimal ann-snn conversion for high-accuracy and ultra-low-latency spiking neural networks.
\newblock {\em arXiv preprint arXiv:2303.04347}, 2023.

\bibitem{hu2023fast}
Yangfan Hu, Qian Zheng, Xudong Jiang, and Gang Pan.
\newblock Fast-snn: Fast spiking neural network by converting quantized ann.
\newblock {\em IEEE Transactions on Pattern Analysis and Machine Intelligence}, 2023.

\bibitem{zhou2022spikformer}
Zhaokun Zhou, Yuesheng Zhu, Chao He, Yaowei Wang, Shuicheng Yan, Yonghong Tian, and Li~Yuan.
\newblock Spikformer: When spiking neural network meets transformer.
\newblock {\em arXiv preprint arXiv:2209.15425}, 2022.

\bibitem{yao2024spike}
Man Yao, Jiakui Hu, Zhaokun Zhou, Li~Yuan, Yonghong Tian, Bo~Xu, and Guoqi Li.
\newblock Spike-driven-transformer.
\newblock {\em Advances in Neural Information Processing Systems}, 36, 2024.

\bibitem{yao2024spike222}
Man Yao, Jiakui Hu, Tianxiang Hu, Yifan Xu, Zhaokun Zhou, Yonghong Tian, Bo~Xu, and Guoqi Li.
\newblock Spike-driven transformer v2: Meta spiking neural network architecture inspiring the design of next-generation neuromorphic chips.
\newblock {\em arXiv preprint arXiv:2404.03663}, 2024.

\bibitem{fang2021incorporating}
Wei Fang, Zhaofei Yu, Yanqi Chen, Timoth{\'e}e Masquelier, Tiejun Huang, and Yonghong Tian.
\newblock Incorporating learnable membrane time constant to enhance learning of spiking neural networks.
\newblock In {\em Proceedings of the IEEE/CVF international conference on computer vision}, pages 2661--2671, 2021.

\bibitem{sironi2018hats}
Amos Sironi, Manuele Brambilla, Nicolas Bourdis, Xavier Lagorce, and Ryad Benosman.
\newblock Hats: Histograms of averaged time surfaces for robust event-based object classification.
\newblock In {\em Proceedings of the IEEE conference on computer vision and pattern recognition}, pages 1731--1740, 2018.

\bibitem{ramesh2019dart}
Bharath Ramesh, Hong Yang, Garrick Orchard, Ngoc~Anh Le~Thi, Shihao Zhang, and Cheng Xiang.
\newblock Dart: distribution aware retinal transform for event-based cameras.
\newblock {\em IEEE transactions on pattern analysis and machine intelligence}, 42(11):2767--2780, 2019.

\bibitem{kim2021optimizing}
Youngeun Kim and Priyadarshini Panda.
\newblock Optimizing deeper spiking neural networks for dynamic vision sensing.
\newblock {\em Neural Networks}, 144:686--698, 2021.

\end{thebibliography}
    
\newpage
\section*{\textbf{APPENDIX}}
\appendix
\section{Detail of Datasets}
\label{sec:detail}

\subsection{Datasets}
This article uses three mainstream neural morphology datasets(DVS128 Gesture, CIFAR10-DVS and N-Caltech 101) and the Imagenet-1K dataset.

\textbf{DVS128 Gesture} is an event-based dataset comprising a sequence of 11 gestures. It consists of 1176 samples in the training set and 288 samples in the test set. The training set was captured by 23 subjects, while the test set was captured by 6 subjects, each under three different lighting conditions. The samples are sized at 128 $\times$ 128 pixels.

\textbf{CIFAR10-DVS} is a conversion from the CIFAR10 dataset, consisting of ten categories totaling 10000 images. The size of each frame has also been expanded to 128 $\times$ 128 pixels. Due to the additional undesired artifacts included in the neuromorphic datasets obtained based on the conversion method, therefore, how to distinguish the categories of targets in complex backgrounds with additional errors is a challenging recognition task.

\textbf{N-Caltech 101} presents a notable challenge, arising from the transformation and slight adaptation of the Caltech dataset to avoid confusion. Comprising 8,709 samples, it is segmented into 100 distinct object classes alongside 1 background class. Emarkably, its image dimensions (180 $\times$ 240) surpass those of numerous other neural morphology datasets.

\textbf{Imagenet-1K} is the most renowned and extensively utilized static image classification dataset. It comprises 1.28 million training samples and 50,000 testing samples, distributed across 1,000 different categories. Each image in the dataset has a resolution of 224x224 pixels.

For the partitioning of datasets like CIFAR10-DVS and N-Caltech101, which necessitate division, we adopt methodologies from prior studies and partition them in a 9:1 ratio. We will furnish the necessary .py scripts for facilitating the partitioning process.

\subsection{Neuromorphic Datasets Preprocessing}
We use the integrating approach to convert event stream for frame data, which is commonly used in SNNs, to preprocess neuromorphic datasets. The coordinate of event can be describe as

\begin{equation}
\label{app-eq1}
\boldsymbol{E}({x_i},{y_i},{p_i})
\end{equation}

where $x_i$ and $y_i$ event's coordinate, $p_i$ represents the event. In order to deduce computational consumption, we group events into $\boldsymbol{T}$ slices, Where $\boldsymbol{T}$ is the network's time simulation step. The process can be describe as

\begin{equation}
\label{app-eq2}
{j_l} = \left\lfloor {\frac{\boldsymbol{N}}{\boldsymbol{T}}} \right\rfloor  \cdot j
\end{equation}

\begin{equation}
\label{app-eq3}
{j_r} = \left\{ {\begin{array}{*{20}{c}}
{\left\lfloor {\frac{\boldsymbol{N}}{\boldsymbol{T}}} \right\rfloor  \cdot (j + 1)}&{j < \boldsymbol{T} - 1}\\
\boldsymbol{N}&{j = \boldsymbol{T} - 1}
\end{array}} \right.
\end{equation}

\begin{equation}
\label{app-eq4}
\boldsymbol{F}(j,p,x,y) = \sum\limits_{i = {j_l}}^{{j_r} - 1} {{\boldsymbol{I}_{p,x,y({p_i},{x_i},{y_i})}}} 
\end{equation}

where $\left\lfloor  \cdot  \right\rfloor $ is the floor operation, and ${{\boldsymbol{I}_{p,x,y({p_i},{x_i},{y_i})}}}$ is an indicator function and it equals 1 only when $(p,x,y) = ({p_i},{x_i},{y_i})$.

\subsection{Data augmentation}

\begin{table}[]
\caption{Data augmentation on CIFAR10-DVS and Imagenet-1K datasets.}
\label{data arg}
\centering
\begin{tabular}{@{} p{2.3cm}<{\centering} @{\hspace{0.2cm}}c@{}}
\toprule
\textbf{Dataset}     & \textbf{Data augmentation}                                                                                                                                       \\ \midrule
CIFAR10-DVS & \begin{tabular}[c]{@{}c@{}}RandomHorizontalFlip(p=0.5)\\ RandomAffine(degree=0, translate=(2.5, 5/128))\end{tabular}                                    \\ \cmidrule(l){2-2} 
Imagenet-1K & \begin{tabular}[c]{@{}c@{}}RandomResizedCrop(224)\\ AutoAugment \\ Normalize(mean={[}0.485, 0.456, 0.406{]}, std={[}0.229, 0.224, 0.225{]})  \\ Mixup (only 104-layers)     \end{tabular} \\ \bottomrule
\end{tabular}
\end{table}

For the DVS128 Gesture and N-Caltech101 datasets, we adhered to our previous experience and refrained from applying data augmentation. To address the significant overfitting issue observed on the CIFAR10-DVS and Imagenet-1K datasets, we implemented data augmentation strategies outlined in Tab.~\ref{data arg}.

\begin{table}[]
\caption{Structures for SMA-ResNet. "CR" and "TR" denote the compression ratios of the C-MSE and T-MSE modules in SMA, respectively.}
\label{resnet_structure}
\centering
\begin{tabular}{|c|c|ccc|}
\hline
Block              & Output Size            & \multicolumn{1}{c|}{SMA-ResNet-18}  & \multicolumn{1}{c|}{SMA-ResNet-34}  & SMA-ResNet-104 \\ \hline
1                  & 112 $\times$ 112                & \multicolumn{3}{c|}{7 $\times$ 7, 64, stride=2}                                                     \\ \hline
\multirow{2}{*}{2} & \multirow{2}{*}{56 $\times$ 56} & \multicolumn{1}{c|}{SMA(CR=4,TR=1)} & \multicolumn{1}{c|}{SMA(CR=4,TR=1)} & SMA(CR=4,TR=1)      \\ \cline{3-5} 
                   &                        & \multicolumn{1}{c|}{$\left[ {\begin{array}{*{20}{c}} {3 \times 3,64}\\ [0pt] {3 \times 3,64} \end{array}} \right] * 2$}               & \multicolumn{1}{c|}{$\left[ {\begin{array}{*{20}{c}} {3 \times 3,64}\\ [0pt] {3 \times 3,64} \end{array}} \right] * 2$}               & $\left[ {\begin{array}{*{20}{c}} {3 \times 3,64}\\ [0pt] {3 \times 3,64} \end{array}} \right] * 2$               \\ \hline
\multirow{2}{*}{3} & \multirow{2}{*}{28 $\times$ 28} & \multicolumn{1}{c|}{SMA(CR=4,TR=1)} & \multicolumn{1}{c|}{SMA(CR=4,TR=1)} & SMA(CR=4,TR=1)      \\ \cline{3-5} 
                   &                        & \multicolumn{1}{c|}{$\left[ {\begin{array}{*{20}{c}} {3 \times 3,128}\\ [0pt] {3 \times 3,128} \end{array}} \right] * 2$}               & \multicolumn{1}{c|}{$\left[ {\begin{array}{*{20}{c}} {3 \times 3,128}\\ [0pt] {3 \times 3,128} \end{array}} \right] * 4$}               & $\left[ {\begin{array}{*{20}{c}} {3 \times 3,128}\\ [0pt] {3 \times 3,128} \end{array}} \right] * 8$               \\ \hline
\multirow{2}{*}{4} & \multirow{2}{*}{14 $\times$ 14} & \multicolumn{1}{c|}{SMA(CR=4,TR=1)} & \multicolumn{1}{c|}{SMA(CR=4,TR=1)} & SMA(CR=4,TR=1)      \\ \cline{3-5} 
                   &                        & \multicolumn{1}{c|}{$\left[ {\begin{array}{*{20}{c}} {3 \times 3,256}\\ [0pt] {3 \times 3,256} \end{array}} \right] * 2$}               & \multicolumn{1}{c|}{$\left[ {\begin{array}{*{20}{c}} {3 \times 3,256}\\ [0pt] {3 \times 3,256} \end{array}} \right] * 6$}               & $\left[ {\begin{array}{*{20}{c}} {3 \times 3,256}\\ [0pt] {3 \times 3,256} \end{array}} \right] * 32$               \\ \hline
\multirow{2}{*}{5} & \multirow{2}{*}{7 $\times$ 7}   & \multicolumn{1}{c|}{SMA(CR=4,TR=1)} & \multicolumn{1}{c|}{SMA(CR=4,TR=1)} & SMA(CR=4,TR=1)      \\ \cline{3-5} 
                   &                        & \multicolumn{1}{c|}{$\left[ {\begin{array}{*{20}{c}} {3 \times 3,512}\\ [0pt] {3 \times 3,512} \end{array}} \right] * 2$}               & \multicolumn{1}{c|}{$\left[ {\begin{array}{*{20}{c}} {3 \times 3,512}\\ [0pt] {3 \times 3,512} \end{array}} \right] * 3$}               & $\left[ {\begin{array}{*{20}{c}} {3 \times 3,512}\\ [0pt] {3 \times 3,512} \end{array}} \right] * 8$               \\ \hline
FC                 & 1 $\times$ 1                    & \multicolumn{3}{c|}{AveragePool, FC(1000)}                                                 \\ \hline
Dropout            & 1 $\times$ 1                    & \multicolumn{2}{c|}{Null}                                                 & Dropout(0.2)   \\ \hline
\end{tabular}
\end{table}

\section{Training details}
\label{sec:model}

\subsection{Network structure}

\begin{table}[]
\caption{Structures for SMA-VGG. "CR" and "TR" denote the compression ratios of the C-MSE and T-MSE modules in SMA, respectively.}
\label{vgg_structure}
\centering
\begin{tabular}{|c|c|ccc|}
\hline
\multirow{2}{*}{Block} & \multirow{2}{*}{SMA-VGG} & \multicolumn{3}{c|}{Output Size}                                                                                                              \\ \cline{3-5} 
                       &                          & \multicolumn{1}{c|}{DVS128 Gesture}                  & \multicolumn{1}{c|}{CIFAR10-DVS}                     & N-Caltech101                    \\ \hline
\multirow{2}{*}{1}     & 3 $\times$ 3,64          & \multicolumn{1}{c|}{128 $\times$ 128}                & \multicolumn{1}{c|}{128 $\times$ 128}                & 180 $\times$ 240                         \\ \cline{2-5} 
                       & MaxPool(2,2,0,1)         & \multicolumn{1}{c|}{64 $\times$ 64}                  & \multicolumn{1}{c|}{64 $\times$ 64}                  & 90 $\times$ 120                          \\ \hline
\multirow{3}{*}{2}     & 3 $\times$ 3,128         & \multicolumn{1}{c|}{\multirow{3}{*}{32 $\times$ 32}} & \multicolumn{1}{c|}{\multirow{3}{*}{32 $\times$ 32}} & \multirow{3}{*}{45 $\times$ 60} \\ \cline{2-2}
                       & SMA(CR=4,TR=4)           & \multicolumn{1}{c|}{}                                & \multicolumn{1}{c|}{}                                &                                 \\ \cline{2-2}
                       & MaxPool(2,2,0,1)         & \multicolumn{1}{c|}{}                                & \multicolumn{1}{c|}{}                                &                                 \\ \hline
\multirow{3}{*}{3}     & 3 $\times$ 3,256         & \multicolumn{1}{c|}{\multirow{3}{*}{16 $\times$ 16}} & \multicolumn{1}{c|}{\multirow{3}{*}{16 $\times$ 16}} & \multirow{3}{*}{22 $\times$ 30} \\ \cline{2-2}
                       & SMA(CR=4,TR=4)           & \multicolumn{1}{c|}{}                                & \multicolumn{1}{c|}{}                                &                                 \\ \cline{2-2}
                       & MaxPool(2,2,0,1)         & \multicolumn{1}{c|}{}                                & \multicolumn{1}{c|}{}                                &                                 \\ \hline
\multirow{3}{*}{4}     & 3 $\times$ 3,512         & \multicolumn{1}{c|}{\multirow{3}{*}{8 $\times$ 8}}   & \multicolumn{1}{c|}{\multirow{3}{*}{8 $\times$ 8}}   & \multirow{3}{*}{11 $\times$ 15} \\ \cline{2-2}
                       & SMA(CR=4,TR=4)           & \multicolumn{1}{c|}{}                                & \multicolumn{1}{c|}{}                                &                                 \\ \cline{2-2}
                       & MaxPool(2,2,0,1)         & \multicolumn{1}{c|}{}                                & \multicolumn{1}{c|}{}                                &                                 \\ \hline
\multirow{3}{*}{5}     & 3 $\times$ 3,512         & \multicolumn{1}{c|}{\multirow{3}{*}{4 $\times$ 4}}   & \multicolumn{1}{c|}{\multirow{3}{*}{4 $\times$ 4}}   & \multirow{3}{*}{5 $\times$ 7}   \\ \cline{2-2}
                       & SMA(CR=4,TR=4)           & \multicolumn{1}{c|}{}                                & \multicolumn{1}{c|}{}                                &                                 \\ \cline{2-2}
                       & MaxPool(2,2,0,1)         & \multicolumn{1}{c|}{}                                & \multicolumn{1}{c|}{}                                &                                 \\ \hline
\multirow{3}{*}{FC-1}  & AveragePool              & \multicolumn{1}{c|}{\multirow{3}{*}{1 $\times$ 1}}   & \multicolumn{1}{c|}{\multirow{3}{*}{1 $\times$ 1}}   & \multirow{3}{*}{1 $\times$ 1}   \\ \cline{2-2}
                       & FC(2048)                 & \multicolumn{1}{c|}{}                                & \multicolumn{1}{c|}{}                                &                                 \\ \cline{2-2}
                       & Dropout(0.5)             & \multicolumn{1}{c|}{}                                & \multicolumn{1}{c|}{}                                &                                 \\ \hline
\multirow{2}{*}{FC-2}  & FC(1024)                 & \multicolumn{1}{c|}{\multirow{2}{*}{1 $\times$ 1}}   & \multicolumn{1}{c|}{\multirow{2}{*}{1 $\times$ 1}}   & \multirow{2}{*}{1 $\times$ 1}   \\ \cline{2-2}
                       & Dropout(0.5)             & \multicolumn{1}{c|}{}                                & \multicolumn{1}{c|}{}                                &                                 \\ \hline
FC-3                   & FC(11/10/101)            & \multicolumn{1}{c|}{1 $\times$ 1}                    & \multicolumn{1}{c|}{1 $\times$ 1}                    & 1 $\times$ 1                    \\ \hline
\end{tabular}
\end{table}

\begin{figure}[ht]
  \centering 
  \includegraphics[width=0.6\textwidth]{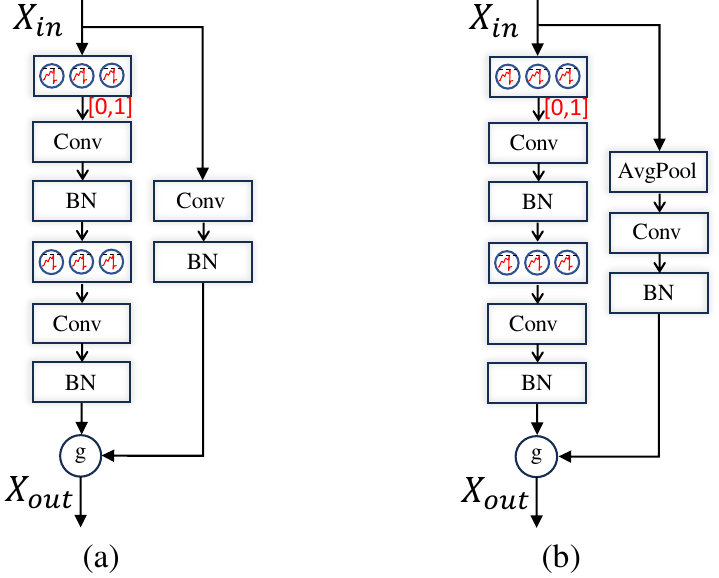}
  \caption{MS ResNet residual block structure. (a) is the structure of residual blocks in MS-ResNet18/34, and (b) is the structure of residual blocks in MS-ResNet104.}
  \label{ms_block} 
\end{figure}

In the experiment detailed in Chap.~4 of the primary text, we employed MS-ResNet networks with varying numbers of layers on the ImageNet-1K dataset, as illustrated in Tab.~\ref{resnet_structure}. Additionally, we utilized an 8-layer VGG architecture on the mainstream neural morphology dataset, as depicted in Tab.~\ref{vgg_structure}. It is worth noting that in both the SMA-ResNet and SMA-VGG architectures, our SMA module closely follows the convolutional layer, and an SMA module is added after each residual connection. This suggests that the actual configuration of the second module in SMA-ResNet18 consists of Conv-SMA-Conv-Conv-SMA-Conv, while the actual structure of the second module of SMA-VGG is Conv-SMA-MaxPool. For clarity, we omitted LIF neurons and BatchNorm in Tab.~\ref{resnet_structure} and~\ref{vgg_structure}. The MS residual block structure in Tab.~\ref{resnet_structure} is detailed in Fig.~\ref{ms_block}, and each convolution operation in Tab.~\ref{vgg_structure} can be interpreted as LIF Neuron-Conv-BN.

\subsection{Training strategy}

\begin{table}[]
\caption{The hyperparameters on each dataset. In the 125-layer structure, we incorporate AZO regularization methods following each SMA, fine-tuning them with hyperparameters as indicated in parentheses.}
\label{exp_hyp}
\centering
\begin{tabular}{@{}cccc@{}}
\toprule
\textbf{Model}                            & \textbf{Dataset}                        & \textbf{Name}           & \textbf{Value}                                                                                  \\ \midrule
\multirow{18}{*}{SMA-VGG}        & \multirow{6}{*}{DVS128 Gesture} & lr             & 1e-4                                                                                   \\
                                 &                                 & T              & 16                                                                                     \\
                                 &                                 & batch\_size    & 9                                                                                     \\
                                 &                                 & train\_epoch   & 200                                                                                    \\
                                 &                                 & loss\_function & MSE\_loss                                                                              \\
                                 &                                 & optimizer      & AdamW(momentum\_decay=1e-3)                                                            \\ \cmidrule(l){2-4} 
                                 & \multirow{6}{*}{CIFAR10-DVS}    & lr             & 1e-3                                                                                   \\
                                 &                                 & T              & 10                                                                                     \\
                                 &                                 & batch\_size    & 24                                                                                     \\
                                 &                                 & train\_epoch   & 200                                                                                    \\
                                 &                                 & loss\_function & MSE\_loss                                                                              \\
                                 &                                 & optimizer      & Adam                                                                                   \\ \cmidrule(l){2-4} 
                                 & \multirow{6}{*}{N-Caltech101}   & lr             & 1e-3                                                                                   \\
                                 &                                 & T              & 14                                                                                     \\
                                 &                                 & batch\_size    & 6                                                                                      \\
                                 &                                 & train\_epoch   & 300                                                                                    \\
                                 &                                 & loss\_function & TET\_loss                                                                              \\
                                 &                                 & optimizer      & NAdam(momentum\_decay=1e-3)                                                            \\ \midrule
\multirow{6}{*}{SMA-ResNet18/34} & \multirow{6}{*}{Imagenet-1K}    & lr             & 0.1                                                                                    \\
                                 &                                 & T              & 6                                                                                      \\
                                 &                                 & batch\_size    & 384/256                                                                                       \\
                                 &                                 & train\_epoch   & 125                                                                                    \\
                                 &                                 & loss\_function & Label\_smoothing(0.1)                                                                  \\
                                 &                                 & optimizer      & \begin{tabular}[c]{@{}c@{}}SGD(Momentum=0.9, \\ weight\_decay=1e-4)\end{tabular}       \\ \midrule
\multirow{6}{*}{SMA-ResNet104}   & \multirow{6}{*}{Imagenet-1K}    & lr             & 0.1(1e-3)                                                                              \\
                                 &                                 & T              & 5                                                                                   \\
                                 &                                 & batch\_size    & 224                                                                              \\
                                 &                                 & train\_epoch   & 125(50)                                                                               \\
                                 &                                 & loss\_function & Label\_smoothing(0.1)                                                                  \\
                                 &                                 & optimizer      & \begin{tabular}[c]{@{}c@{}}SGD(Momentum=0.9, \\ weight\_decay=1e-4(1e-5)\end{tabular} \\ \bottomrule
\end{tabular}
\end{table}

\begin{table}[]
\caption{Neuronal configuration parameters of LIF.}
\label{lif_neuron}
\centering
\begin{tabular}{@{}ccc@{}}
\toprule
\textbf{Datasets}                               & \textbf{Parameter}                      & \textbf{Value}                                                                         \\ \midrule
\multirow{4}{*}{Neuromorphic Datasets} & ${\boldsymbol{U}_{threshold}}$ & 1.0                                                                           \\
                                       & ${\boldsymbol{U}_{reset}}$     & 0.0                                                                           \\
                                       & $\tau$                         & 2.0                                                                           \\
                                       & Surrogate Gradient Function    & ATan()                                                                        \\ \midrule
\multirow{4}{*}{Imagenet-1K}           & ${\boldsymbol{U}_{threshold}}$ & 0.5                                                                           \\
                                       & ${\boldsymbol{U}_{reset}}$     & 0.0                                                                           \\
                                       & $\tau$                         & 4.0                                                                           \\
                                       & Surrogate Gradient Function    & $ sign(\left| {\boldsymbol{I}_t^n - {\boldsymbol{U}_{threshold}}} \right| \le \frac{1}{2}) $ \\ \bottomrule
\end{tabular}
\end{table}

The experiments conducted on the neuromorphic datasets in this article were all carried out using 4 $\times$ 3090GPUs, while experiments on Imagenet-1K were performed using 6/8/10 $\times$ 3090D GPUs depending on the ResNet layers. Tab.~\ref{exp_hyp} presents the hyperparameters necessary for training each dataset, while Tab.~\ref{lif_neuron} delineates the membrane constants of LIF neurons for both the neuromorphic dataset and Imagenet-1K dataset.

\subsection{Equipment details}

\begin{table}[]
\caption{Equipment configuration used in the experiment.}
\label{equipment}
\centering
\begin{tabular}{@{}ccccccc@{}}
\toprule
No & CPU           & GPU            & Memory & CUDA & Pytorch & OS           \\ \midrule
1         & Gold 6133 $ \times $ 2 & RTX 3090 $ \times $ 4   & 128GB  & 12.1 & 2.0.1   & Ubuntu 18.04 \\
2         & Gold 6430     & RTX 4090 $ \times $ 8   & 960GB  & 12.1 & 2.1.0   & Ubuntu 22.04 \\
3         & AMD EPYC 9754 & RTX 4090D $ \times $ 10 & 600GB  & 12.1 & 2.1.0   & Ubuntu 22.04 \\ \bottomrule
\end{tabular}
\end{table}

This research involves three experiments, the configurations of which are detailed in Tab.~\ref{equipment}. Both Equipment 2 and Equipment 3 were leased from the \href{https://www.autodl.com}{AutoDL} cloud computing platform.

\section{Hyperparameter selection}
\label{sec:hyperp}

For general attention modules, particularly those based on SE modules, the hyperparameters within the modules are crucial, as they frequently have the potential to influence the performance of attention mechanisms to a certain extent.

\subsection{SMA hyperparameter selection}

\begin{figure}[ht]
  \centering 
  \includegraphics[width=1\textwidth]{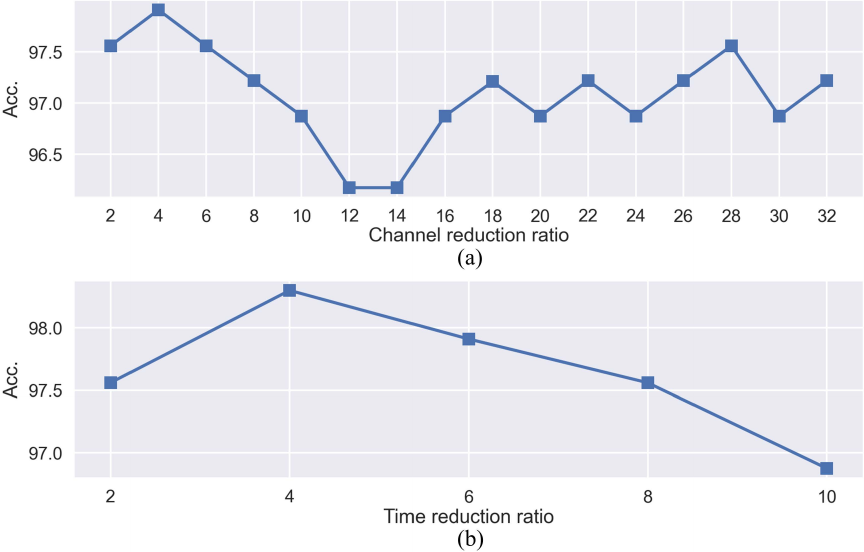}
  \caption{Explore the best reduction ratio. The accuracy of the VGG network without any attention added is 97.92\% . (a) Only channel dimension attention has been added in. (b) Explored the time dimension reduction ratio based on a channel reduction ratio of 4.}
  \label{sma_hyp} 
\end{figure}

The SMA module features only two hyperparameters: the channel reduction ratio and the time reduction channel. As in Chap.~4 of the primary text, we undertook experiments using the DVS128 Gesture dataset to ascertain the optimal reduction ratios in both the channel and time dimensions. Initially, we integrated SMA solely along the channel dimension into the network for experimentation, with the results depicted in Fig.~\ref{sma_hyp}(a). Notably, setting the reduction ratio in the channel dimension to 4 yielded the most favorable outcomes. Building upon this finding, we further investigated the optimal reduction ratio in the time dimension, as illustrated in Fig.~\ref{sma_hyp}(b), where the optimal setting remained at 4.

The insights from Fig.~\ref{sma_hyp} highlight the critical role of hyperparameter selection within the attention module, as an incorrect choice can detrimentally impact model performance. Upon close examination of the findings in Fig.~\ref{sma_hyp}, it becomes evident that excessive compression may blur the importance of information within the attention module, while insufficient compression may prevent the model from achieving the correct balance of importance.

\subsection{AZO hyperparameter selection}
\label{sec:AZO Hyp}

\begin{figure}[ht]
  \centering 
  \includegraphics[width=1\textwidth]{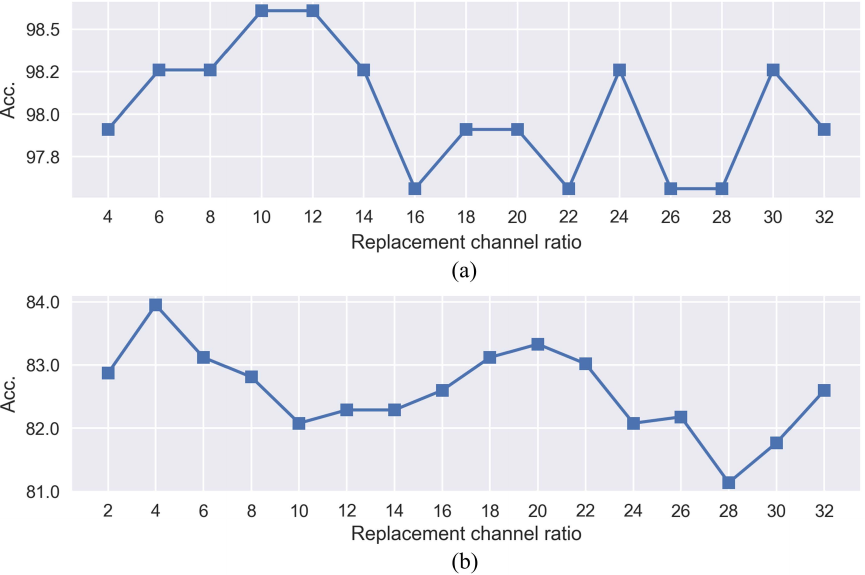}
  \caption{Explore the best RCR. The RTR for all experiments is 4. (a) is an experiment on the DVS128 Gesture dataset, and (b) is an experiment on the CIFAR10-DVS dataset.}
  \label{azo_hyp} 
\end{figure}

\begin{figure}[ht]
  \centering 
  \includegraphics[width=1\textwidth]{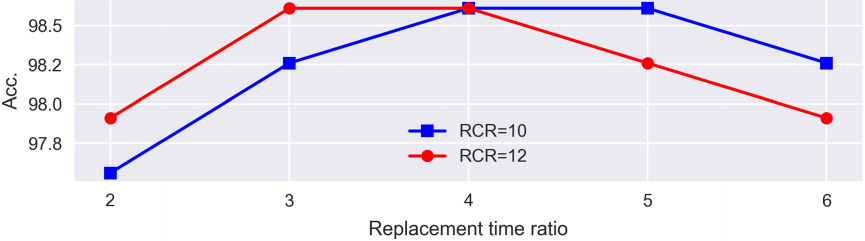}
  \caption{Explore the best RTR while the RCR has been determined.}
  \label{azo_hyp_t} 
\end{figure}

The AZO regularization method utilizes two hyperparameters: the Replacement Time Ratio(RTR) and the Replacement Channel Ratio(RCR). Given the short simulation timesteps in our experiment, the influence of the timestep on the AZO effect appears to be considerably less significant than the number of channels, a notion supported by Fig.~\ref{azo_hyp_t}. Given the diverse nature of the datasets involved—CIFAR10-DVS and N-Caltech101, which are converted datasets, and DVS128 Gesture, which represents natural neural morphology—we investigated the optimal RCR for the DVS128 Gesture and CIFAR10-DVS datasets, setting the RTR at 4, as illustrated in Fig.~\ref{azo_hyp}. Following this, we further explored the optimal RTR for the DVS128 Gesture dataset, as depicted in Fig.~\ref{azo_hyp_t}.

\begin{table}[]
\caption{The best settings for AZO.}
\label{azo_setting}
\centering
\begin{tabular}{@{}cccc@{}}
\toprule
\textbf{AZO hyperparameters}            & \textbf{DVS128 Gesture} & \textbf{CIFAR10-DVS} & \textbf{N-Caltech101} \\ \midrule
Best RCR & 10/12             & 4           & 24           \\
Best RTR    & 4              & 4           & 4            \\ \bottomrule
\end{tabular}
\end{table}

We posit that the distinct data formats and acquisition methods necessitate uniquely suitable hyperparameter settings for each dataset when using the AZO regularization method. Due to the extensive data volume, we refrained from conducting an exhaustive exploration of optimal hyperparameters on the N-Caltech101 dataset, opting instead to provide a set of parameters that demonstrated improved performance, as detailed in Tab.~\ref{azo_setting}.

\subsection{Comparison of accuracy between LIF based and ReLU based SMA}
\label{sec:comparison}

\begin{table}[]
\caption{Comparing the effects of LIF neurons and ReLU in SMA}
\label{lif_and_relu}
\centering
\begin{tabular}{@{}ccccccc@{}}
\toprule
\multirow{2}{*}{\textbf{Type}} & \multicolumn{2}{c}{\textbf{DVS128 Gesture}} & \multicolumn{2}{c}{\textbf{CIFAR10-DVS}} & \multicolumn{2}{c}{\textbf{N-Caltech101}} \\ \cmidrule(l){2-7} 
                      & SMA-C            & SMA             & SMA-C           & SMA           & SMA-C           & SMA            \\ \midrule
ReLU                  & 97.9             & 98.3            & 82.3            & 83.1          & 82.7           & 83.7           \\
LIF Neuron            & 97.9             & 98.3            & 81.8            & 82.6          & 82.5           & 82.9          \\ \bottomrule
\end{tabular}
\end{table}

In the Discussion section of the primary text, we conducted extensive experiments utilizing the LIF version of the SMA module. The discrete nature of signals in SNNs facilitated easy visualization and analysis of attention modules based on spike count or firing rate—a method widely adopted in numerous studies. As illustrated in Tab.\ref{lif_and_relu}, our experiments showcased that while the ReLU version of the SMA module exhibits slightly superior performance, the disparity in performance between it and the LIF version is minimal. Notably, even on the DVS128 Gesture dataset, both versions yielded identical effects. Hence, it is reasonable to posit that the LIF version and the ReLU version of the SMA module share the same attention mechanism, affirming our discussion in the primary text.

\section{Further discussion}
\label{sec:further}

\subsection{Complexity}
\label{sec:further:comp}
For an efficient plug-and-play attention module, it's crucial to conduct a detailed analysis of both time and space complexities. In this chapter, we specifically concentrate on the process involved in computing attention weights, setting aside the Multiscale Coding process. The latter is primarily composed of straightforward convolution and addition operations, which are relatively less complex. For Channel Multiscale Attention and Time Multiscale Attention in SMA. In section 3.2 of the primary text, we describe them as follows:

\begin{equation}
\label{app-eq5}
f_\alpha ^n(\boldsymbol{X}) = \boldsymbol{C}onv2d(\boldsymbol{K}_{E,\alpha }^n,\delta (\boldsymbol{C}onv2d({\boldsymbol{K}_{S,\alpha }},\boldsymbol{X}))),
\end{equation}

\begin{equation}
\label{app-eq6}
f_\beta ^n(\boldsymbol{X}) = \boldsymbol{C}onv2d(\boldsymbol{K}_{E,\beta }^n,\delta (\boldsymbol{C}onv2d({\boldsymbol{K}_{S,\beta }},\boldsymbol{X}))),
\end{equation}

\begin{equation}
\label{app-eq7}
\begin{array}{*{20}{c}}
{{\boldsymbol{W}_\alpha } = \boldsymbol{S}oft\boldsymbol{M}ax({f_\alpha }(\boldsymbol{A}vgpool({\boldsymbol{Y}_t}))),}&{{\boldsymbol{W}_\alpha } \in {\mathbb{R}^{N \times T \times 1}}}
\end{array}
\end{equation}

\begin{equation}
\label{app-eq8}
\begin{array}{*{20}{c}}
{{\boldsymbol{W}_{\beta ,t}} = \boldsymbol{S}oft\boldsymbol{M}ax({f_\beta }({\boldsymbol{Y}_t})),}&{{\boldsymbol{W}_{\beta ,t}} \in {\mathbb{R}^{N \times C \times 1}}}
\end{array}
\end{equation}

\begin{table}[]
\caption{Inference time.}
\label{inference_time}
\centering
\begin{tabular}{@{}cccc@{}}
\toprule
\multirow{2}{*}{\textbf{Dataset}}     & \multirow{2}{*}{\textbf{Model}} & \multicolumn{2}{c}{\textbf{Inference time(s)}} \\ \cmidrule(l){3-4} 
                             &                        & \textbf{No SMA}        & \textbf{With SMA}              \\ \midrule
DVS128 Gesture               & VGG                    & 2.10          & 2.97       \\
CIFAR10-DVS                  & VGG                    & 3.26          & 4.79     \\
N-Caltech101                 & VGG                    & 7.29          & 10.91    \\
\multirow{3}{*}{Imagenet-1K} & ResNet18               & 14.78         & 25.41                      \\
                             & ResNet34               & 19.34         & 29.87                      \\
                             & ResNet104              & 39.99         & 52.25                     \\ \bottomrule
\end{tabular}
\end{table}


\textbf{Time complexity}. In analyzing the Time Multiscale Attention within SMA for multiscale encoded data sized $ \boldsymbol{C} \times \boldsymbol{T} \times 1 $, the Avgpool operation has a time complexity of $\boldsymbol{o}(\boldsymbol{C} \times \boldsymbol{T})$. Post-Avgpool, the data size is reduced to $\boldsymbol{T} \times 1$. For this input data, the T-MSE Block initiates with a $1 \times 1 $ convolution operation, which has a time complexity of $ \boldsymbol{o}({\boldsymbol{T}^2}) $. It then proceeds to filter out negative values using the ReLU function, also with a time complexity of $ \boldsymbol{o}(\boldsymbol{T}) $. Following this, another $ 1 \times 1 $ convolution is performed at various scales, with its own time complexity of $ \boldsymbol{o}(\boldsymbol{N} \times {\boldsymbol{T}^2}) $. For processing the output results from the T-MSE Block, a SoftMax operation is necessary, adding a further time complexity of $ \boldsymbol{o}(\boldsymbol{N} \times \boldsymbol{T}) $. Consequently, the total time complexity for the Time Multiscale Attention module sums up to $ \boldsymbol{o}(\boldsymbol{N} \times {\boldsymbol{T}^2}) $.

Channel Multiscale Attention operates similarly to Time Multiscale Attention but without the Avgpool process. Instead, attention weights are computed at each timestep through the C-MSE Block. Consequently, the overall time complexity for Channel Multiscale Attention is $ \boldsymbol{o}(\boldsymbol{T} \times \boldsymbol{N} \times {\boldsymbol{C}^2}) $. Given that $ \boldsymbol{T} $ and $ \boldsymbol{N} $ often represent smaller values, SMA only slightly increases inference time, and its additional inference time is independent of the depth of the network, as demonstrated in Tab.~\ref{inference_time}.

\textbf{Spatial complexity}. When calculating attention weights, the sole additional computational units that introduce parameters are the 1x1 convolutional layers in the T-MSE and C-MSE blocks. The spatial complexity for T-MSE is $ \boldsymbol{o}({\boldsymbol{N} \times {\boldsymbol{T}^2}}) $, and for C-MSE, it is $ \boldsymbol{o}(\boldsymbol{T} \times \boldsymbol{N} \times {\boldsymbol{C}^2}) $. The additional parameters required by the convolution operation in multiscale encoders are unavoidable. However, the attention weight calculation (decoder) introduces only a minimal number of parameters.

\subsection{Optimizing the implementation of AZO}

\begin{table}[]
\caption{Comparison of Efficiency between Tensor Parallel and Explicit Loop Implemented AZO Regularization Methods.}
\label{AZO time}
\centering
\begin{tabular}{@{}cccc@{}}
\toprule
\multirow{2}{*}{Version} & \multicolumn{3}{c}{Training time(s)}           \\ \cmidrule(l){2-4} 
                         & Dvs128 Gesture & CIFAR10-DVS & N-Caltech101 \\ \midrule
Explicit Loop            & 55.29          & 802.54      & 755.07             \\
Tensor Parallelism       & 4.35           & 68.61       & 44.04        \\ \bottomrule
\end{tabular}
\end{table}

Clearly, the efficiency of implementing AZO regularization in Python using nested display loops is quite low. Thus, we propose a method for AZO implementation that leverages the tensor parallelism of Python and Numpy. As illustrated in Tab.~\ref{AZO time}, we compared the time required for a single training iteration between the parallel and the loop-based versions of AZO. We found that the loop version offers a significant efficiency advantage, establishing a solid foundation for the widespread adoption of AZO-like regularization algorithms.

\subsection{AZO effect}

\begin{figure}[ht]
  \centering 
  \includegraphics[width=1\textwidth]{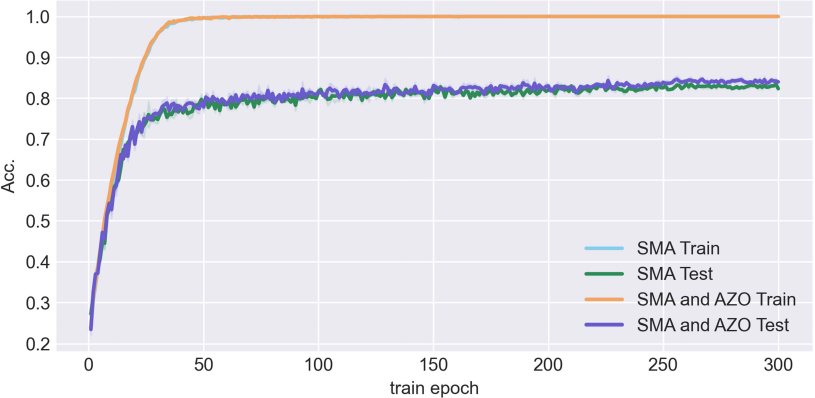}
  \caption{Comparison of generalization errors for five runs.}
  \label{AZO_Ncaltech} 
\end{figure}

As elucidated in Section 3.3 of the primary text, the incorporation of the AZO algorithm is posited to enhance the generalizability of the model. This postulate is grounded in the algorithm's methodology, which perturbs the hidden unit activations with noise during the training of pseudo-sets. To empirically substantiate this hypothesis, we employed the N-Caltech101 dataset—a dataset of considerable complexity, meticulously selected from among a consortium of smaller datasets. Fig.~\ref{AZO_Ncaltech} provides a graphical representation of the empirical results, contrasting the model’s performance on training and testing accuracies both pre and post the integration of the AZO algorithm. The results unambiguously indicate that the AZO algorithm materially amplifies the generalization capabilities of the model.

\subsection{The effect of SMA and AZO on ResNet Structure}

\begin{table}[]
\caption{The efficacy of SMA and AZO on small-scale datasets. We are still using the MS ResNet network architecture.}
\label{resnet_mini}
\centering
\begin{tabular}{@{}ccc@{}}
\toprule
Work             & DVS128 Gesture & CIFAR10-DVS \\ \midrule
SMA-ResNet18     & +0.69\%        & +3.46\%     \\
SMA-AZO-ResNet18 & +1.04\%        & +4.96\%     \\ \bottomrule
\end{tabular}
\end{table}

In the main text, we exclusively showcased the efficacy of SMA within the ResNet architecture on extensive datasets such as Imagenet-1K. However, in this section, we extend our analysis to demonstrate its impact on smaller-scale datasets. Tab.~\ref{resnet_mini} illustrates that SMA yields notable improvements within the ResNet framework even for datasets of reduced scale.

\end{document}